\documentclass[lettersize,journal]{IEEEtran}
\usepackage{amsmath,amsfonts,amssymb}
\usepackage{balance}
\usepackage{algorithmic}
\usepackage{algorithm}
\usepackage[ruled,vlined,linesnumbered,algo2e]{algorithm2e}
\usepackage{array}
\usepackage[dvipsnames]{xcolor}
\usepackage{multirow}
\usepackage{pifont}
\usepackage{nccmath} 
\usepackage{gensymb} 
\usepackage[caption=false,font=normalsize,labelfont=sf,textfont=sf]{subfig}
\usepackage{textcomp}
\usepackage{stfloats}
\usepackage{arydshln}
\usepackage{url}
\usepackage{hyperref}

\usepackage{verbatim}
\usepackage{graphicx}
\usepackage{cite}
\usepackage{bm} 
\usepackage{booktabs}
\newtheorem{theorem}{Theorem}[section]
\newtheorem{remark}[theorem]{Remark}

\newcommand{\Psib}{\mathbf{\Psi}}

\newcommand{\real}{\mathbb{R}}
\newcommand{\realplus}{\real_{\geq 0}}
\newcommand{\cplx}{\mathbb{C}}

\newcommand{\range}{\mathcal{R}}

\newcommand{\Ac}{\mathcal{A}}

\newcommand{\Fc}{\mathcal{F}}
\newcommand{\Gc}{\mathcal{G}}
\newcommand{\Hc}{\mathcal{H}}

\newcommand{\Kc}{\mathcal{K}}
\newcommand{\Lc}{\mathcal{L}}

\newcommand{\Pc}{\mathcal{P}}

\newcommand{\Sc}{\mathcal{S}}
\newcommand{\Tc}{\mathcal{T}}
\newcommand{\Uc}{\mathcal{U}}

\newcommand{\Xc}{\mathcal{X}}

\newcommand{\Phib}{\boldsymbol{\Phi}}

\newcommand{\Pf}{\mathfrak{P}}

\newcommand{\Kedmd}{K_{\operatorname{EDMD}}}

\newcommand{\until}[1]{\{1,\dots,#1\}}

\newcommand{\Span}{\operatorname{span}}

\newcommand{\edmd}{\operatorname{EDMD}}

\newcommand{\ic}{\mathcal{I}_C}

\newcommand{\restr}[2]{#1 \!\! \restriction_{#2}}

\newcommand{\longthmtitle}[1]{\mbox{}{\textit{(#1):}}}

\newcommand{\oprocendsymbol}{\hbox{$\square$}}
\newcommand{\oprocend}{\relax\ifmmode\else\unskip\hfill\fi\oprocendsymbol}

\makeatletter
\newcommand\fs@spaceruled{\def\@fs@cfont{\bfseries}\let\@fs@capt\floatc@ruled
  \def\@fs@pre{\vspace{1.2\baselineskip}\hrule height.8pt depth0pt \kern2pt}%
  \def\@fs@post{\kern2pt\hrule\kern-12pt}%
  \def\@fs@mid{\kern2pt\hrule\kern2pt}%
  \let\@fs@iftopcapt\iftrue}
\makeatother

\begin{document}

\title{Koopman Operators in Robot Learning}

\author{Lu Shi,$^{1,7}$ Masih Haseli,$^{2}$ Giorgos Mamakoukas,$^{3}$ Daniel Bruder,$^{4}$ Ian Abraham,$^{5}$\\ Todd Murphey,$^{6}$ Jorge Cort\'es,$^{2}$ and Konstantinos Karydis$^{1}$~\IEEEmembership{}
\thanks{$^{1}$Dept. of Electrical and Computer Engineering, University of California, Riverside, $^{2}$Dept. of Mechanical and Aerospace Engineering,
University of California San Diego, $^{3}$Dept. of Planning and Control, Zoox Inc., $^{4}$Dept. of Mechanical Engineering, University of Michigan, $^{5}$Dept. of Mechanical Engineering, Yale University, $^{6}$Dept. of Mechanical Engineering, Northwestern University, $^{7}$Institute of AI Industry Research (AIR), Tsinghua University. 
Emails: shilu@air.tsinghua.edu.cn, kkarydis@ece.ucr.edu, \{mhaseli, cortes\}@ucsd.edu, t-murphey@northwestern.edu, ian.abraham@yale.edu, bruderd@umich.edu, giorgosmamakoukas@u.northwestern.edu.}
\thanks{
We gratefully acknowledge the support of NSF under grants \{IIS-1910087, CMMI-2133084, CNS-2237576, and CMMI-2046270\}, of ONR under grants \{N00014-19-1-2264, N00014-23-1-2353, and N00014-21-1-2706\}, and a Shuimu Scholarship of Tsinghua University. Any opinions, findings, and conclusions or recommendations expressed in this material are those of the authors and do not necessarily reflect the views of the funding agencies.}
\thanks{}}

\markboth{Koopman Operators in Robot Learning}{}

\maketitle

\begin{abstract}
Koopman operator theory offers a rigorous treatment of dynamics and has been emerging as an alternative modeling and learning-based control method across various robotics sub-domains. 
Due to its ability to represent nonlinear dynamics as a linear (but higher-dimensional) operator, Koopman theory offers a fresh lens through which to understand and tackle the modeling and control of complex robotic systems. 
Moreover, it enables incremental updates and is computationally inexpensive, thus making it particularly appealing for real-time applications and online active learning. 
This review delves deeply into the foundations of Koopman operator theory and systematically builds a bridge from theoretical principles to practical robotic applications. 
We begin by explaining the mathematical underpinnings of the Koopman framework and discussing approximation approaches for incorporating inputs into Koopman-based modeling. Foundational considerations, such as data collection strategies as well as the design of lifting functions for effective system embedding, are also discussed. 
We then explore how Koopman-based models serve as a unifying tool for a range of robotics tasks, including model-based control, real-time state estimation, and motion planning. 
The review proceeds to a survey of cutting-edge research that demonstrates the versatility and growing impact of Koopman methods across diverse robotics sub-domains: from aerial and legged platforms to manipulators, soft-bodied systems, and multi-agent networks. 
A presentation of more advanced theoretical topics, necessary to push forward the overall framework, is included. 
Finally, we reflect on some key open challenges that remain and articulate future research directions that will shape the next phase of Koopman-inspired robotics. To support practical adoption, we provide a hands-on tutorial with executable code at \url{https://shorturl.at/ouE59}.
\end{abstract}


\begin{IEEEkeywords}
Koopman operator theory, robotics, modeling, control, learning.
\end{IEEEkeywords}

\section{Introduction}\label{sec:Intro}
\IEEEPARstart{R}{untime} learning is an open challenge in robotics. 
The wealth of robot learning methods being actively investigated (for example, neural ODEs~\cite{chen2018neuralODE}, [deep] reinforcement learning~\cite{kober2013RLROBO}, and generative AI~\cite{mandapuram2018GAI}) all rely on large amounts of data collected in an offline manner, mostly via high-fidelity simulators. 
However, real-world deployment requires the robots to operate in environments that are often \emph{novel}, characterized by phenomena not anticipated by \emph{a priori} datasets, and incorporating interactions that are \emph{unsimulable}. 
These three foundational characteristics transcend robotics applications. For example, human-robot interaction (HRI)~\cite{sheridan2016HRI} may take place in an engineered environment, but it may be impossible to model people based on first-principle statements or average characteristics such as those obtained from large datasets. 
Similarly, wholly physical systems---e.g., soft matter~\cite{lee2017soft}, systems in turbulent flow~\cite{christensen2022underwater, karydis2015probabilistically}, mechanoreception of touch~\cite{lee2020CR}---may not be meaningfully ``simulatable" based on first principles with current technology, either because the complexity of such a simulation is impractical (e.g., necessitating to solve complex partial differential equations [e.g., Navier-Stokes]) or because the system has both unknown and unknowable parameters and boundary conditions (e.g., predictive mechanics of mechanoreception in haptics); in some cases, first principles may not even be readily available (e.g., robots interacting with people with unknown intent). Other cases include wearable technologies~\cite{godfrey2018WT,kokkoni2020development,mucchiani2022closed}, and robotics in unstructured and dynamic environments~\cite{aoude2013uncertain, kan2020online}. 

The emphasis in the last decade on deep learning has contributed to reliance on `big data' in robotics~\cite{levine2018learning, levine2016end}, driven by slow algorithms operating on offline data that have been labeled at scale (e.g., the massive availability of labeled images, contributing to the prioritization of vision as a sensing modality in robotics). In contrast, our effort herein centers on the following motivating question. \emph{If robots---with their physical bodies and embodied learning constraints---are going to function in novel settings without highly relying on offline data, what tools are available that are appropriate for runtime learning using only `small data?'}   

\begin{figure*}[t]
\vspace{0pt}
    \centering
         \includegraphics[width = 0.76\textwidth]{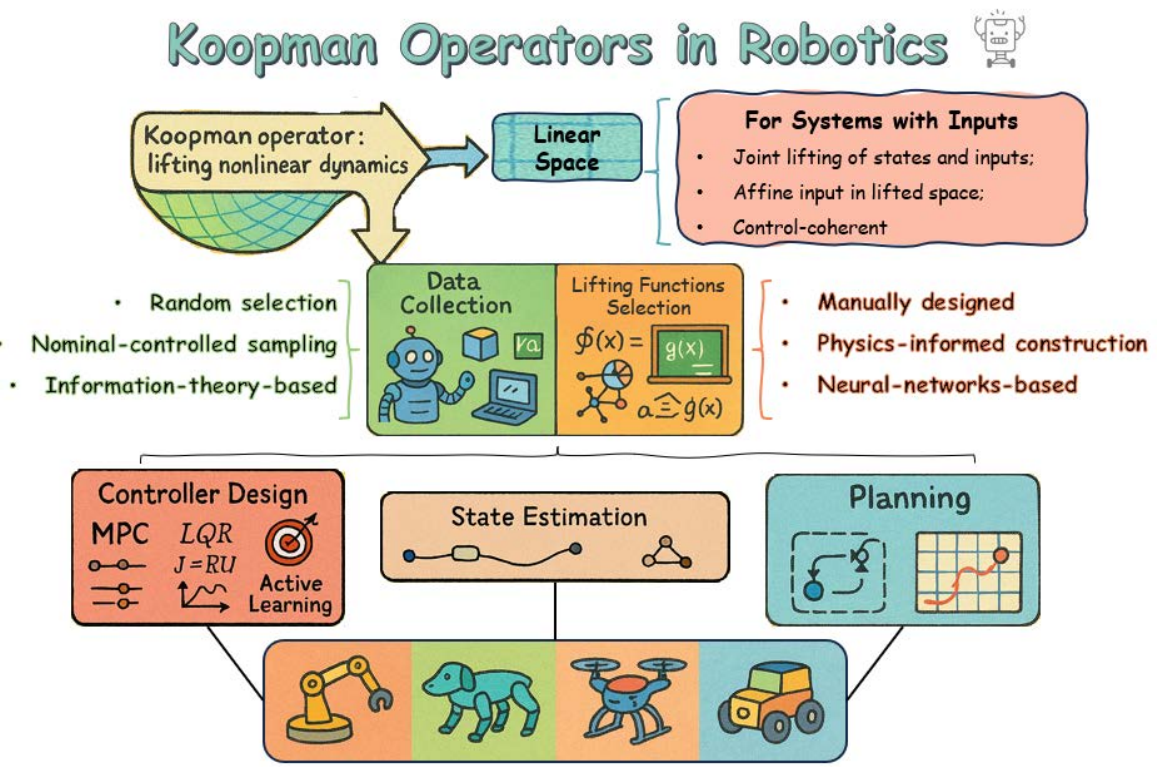}
    \vspace{-6pt}
    \caption{Overview of the utility and integration of Koopman operator theory and practice in robotics.}
    \label{fig:overview}
    \vspace{-9pt}
\end{figure*}

Koopman operators provide a partial answer to this question. 
The main way they are integrated within core robotics foundational research (modeling, control and estimation, and motion planning) is summarized in Fig.~\ref{fig:overview}. 
The essential ingredients in Koopman operators are an infinite basis of observables, a linear operator predicting the evolution of those observables, and some method for approximating the operator using finite calculations. 
These ingredients are sufficient to represent any nonlinear system, including nonsmooth and discontinuous ones~\cite{asada2023global}. 
These operators have only recently made their way into the field of robotics, both theoretically and experimentally. 
Machine learning (data-driven) techniques using Koopman operators are suited to robotics needs---both in terms of \textbf{empirical properties} and \textbf{formal properties}. 
Empirically, Koopman operators only require sparse data sets, making them amenable to runtime computation using only small datasets~\cite{zhang2019online}. 
Owing to their runtime computation affordances, Koopman operators can facilitate adaptive system behavior (e.g., spontaneous control response to unmodeled turbulent flow), that is hard to achieve otherwise.  
Formal properties include applying physically relevant properties to otherwise unknown models (e.g., stability, invariance, or symmetry), the ability to directly calculate information measures for continuous active learning, the benefit of applying linear control techniques (e.g., LQR) to nonlinear systems and the ease of certifying stability (e.g., constructive Control Lyapunov Functions~\cite{deka2022Lyapunov}), and guarantees on sample efficiency. 
These benefits collectively contribute data-driven techniques that can be implemented synchronously with robotic execution, rather than in an offline batch processing manner.


Koopman operators can offer a unified framework for modeling nonlinear dynamics by providing more efficient and versatile representations of robot behavior, which in turn can facilitate reactive motion planning and adaptive control. 
Such models have the following three distinct advantages. 
(1) \textbf{Interpretability}: Unlike many machine learning approaches that represent the input-output relationship as a black box neural network (NN), the Koopman operator theory provides dynamical model descriptions of the underlying system rooted at principled geometric and algebraic properties that can be leveraged to explain the performance of data-driven approximations.  
(2) \textbf{Data-efficiency}: The Koopman operator theory demands only a limited number of measurements when compared to most of the NN-based methods, making it suitable for real-time implementation. 
(3) \textbf{Linear Representation}: The Koopman operator leads to a linear representation of the dynamics, enabling the use of linear systems tools and methods for analysis, estimation, and control synthesis.    




Koopman operator theory applied to robotics is relatively new (the bulk of works discussed herein have appeared within the past five years) and continues to rapidly evolve to introduce algorithmic enhancements and validation techniques for modeling and `downstream' tasks such as controller design (Fig.~\ref{fig:overview}). 
While existing review articles have extensively discussed data-driven methodologies~\cite{brunton2019DDReview} and operator-based algorithms~\cite{kaiser2020operatorReview}, recent review articles on Koopman operator theory have predominantly focused on theoretical analyses~\cite{otto2021koopmanReview,bevanda2021koopmanReview,brunton2022koopmanReview}, encompassing topics such as operator estimation, spectral properties analysis, and controller design~\cite{taylor202ALreview}. 
Further, a recent review focused exclusively on soft robotics~\cite{shi2023review}.  

Here, our goal is to offer a comprehensive introduction to the Koopman operator and how it applies to robotics with a range of specific examples across various robotics sub-domains. 
We seek to highlight the benefits of Koopman-based approaches for modeling, control and estimation, and motion planning. 
Additionally, we seek to consolidate and highlight recent trends in research efforts related to the practical utilization of Koopman operator theory across various robotic systems. 
To this end, the review is organized as follows. 
We begin in Section~\ref{Sec:Fundamental} with the fundamentals of Koopman operator theory, where we introduce the core mathematical concepts, data-driven estimation, and prediction methods. Section~\ref{sec:model} transitions into the use of Koopman-based models in robotics and their usage for control, state estimation, and planning. 
Section~\ref{sec:Robots} presents a system-level breakdown of Koopman applications across robotics sub-domains (including manipulation, legged locomotion, soft and continuum robots, aerial and underwater vehicles, and multi-agent systems). 
To strengthen the ties to control theory and summarize the important theoretical underpinnings that are required to push forward the state of the underlying theory, Section~\ref{sec:advanced} presents more advanced topics on Koopman-based modeling and control for general (nonlinear) systems. 
Lastly, Section~\ref{sec:Discussion} outlines future research directions and Section~\ref{sec:Conclusion} concludes the review.

\section{Fundamentals of Koopman Operator Theory}\label{Sec:Fundamental}
In 1931, Bernard O. Koopman~\cite{koopman1931koopman} demonstrated that a nonlinear dynamical system can be represented using an infinite-dimensional linear operator acting on a Hilbert space of measurement functions of the system's state.  
Owing to this, the Koopman operator has recently gained attention in robotics for various tasks, such as modeling, control, planning, and observer design. 
In this section, we introduce the fundamentals of Koopman Operator Theory. 
We begin by summarizing the core mathematical concepts,\footnote{~A more extensive and rigorous treatise is provided in Section~\ref{sec:advanced}.} 
and continue to present how to estimate the operator directly from data. 
Since the Koopman operator was originally formulated for unforced systems, 
we also explain how Koopman approximations can be extended to systems with inputs. 
Key notation is summarized in Table~\ref{tab:notation}.
\vspace{-12pt}
\begin{table}[h]
\caption{\textbf{Key Notation used in the review.}}
\label{tab:notation}
\vspace{-9pt}
\begin{center}
\begin{tabular}{l|l}
\toprule
Notation & Description  \\
\midrule
$x$ & State\\
\midrule
$u$ & Input\\
\midrule
$T$ & System's original (nonlinear) propagation rule\\
\midrule
$N_{x}$& Dimension of state $x$ \\
\midrule
$N_u$& Dimension of input $u$  \\
\midrule
$t$& Index for online system propagation  \\
\midrule
$g$ & Observable function\\
\midrule
$\mathcal{K}$ & Koopman operator of discrete-time system\\
$K$ & Approximated finite-dimensional Koopman operator\\
\midrule
$\Pf_{\Kc f}^{\edmd}$ & EDMD Predictor\\
\midrule
$\phi$ & Koopman eigenfunction\\
$\lambda_{\phi}$ & Koopman eigenvalue w.r.t $\phi$\\ 
$v_{\phi}$ & Left eigenvector of $\Kedmd$\\
\midrule
\multirow{2}{*}{$N$}& Dimension of observables' dictionary;\\
& Size of estimated Koopman operator \\
\midrule
$\psi_n$ & The $n$-th lifting function\\
$\mathbf{\Psi}$ & Vector-valued observable function\\
\bottomrule
\end{tabular}
\end{center}
\vspace{-24pt}
\end{table}

\subsection{Koopman Operator Fundamentals}
\begin{figure}[!t]
\vspace{6pt}
    \centering
    \includegraphics[width = 0.47\textwidth]{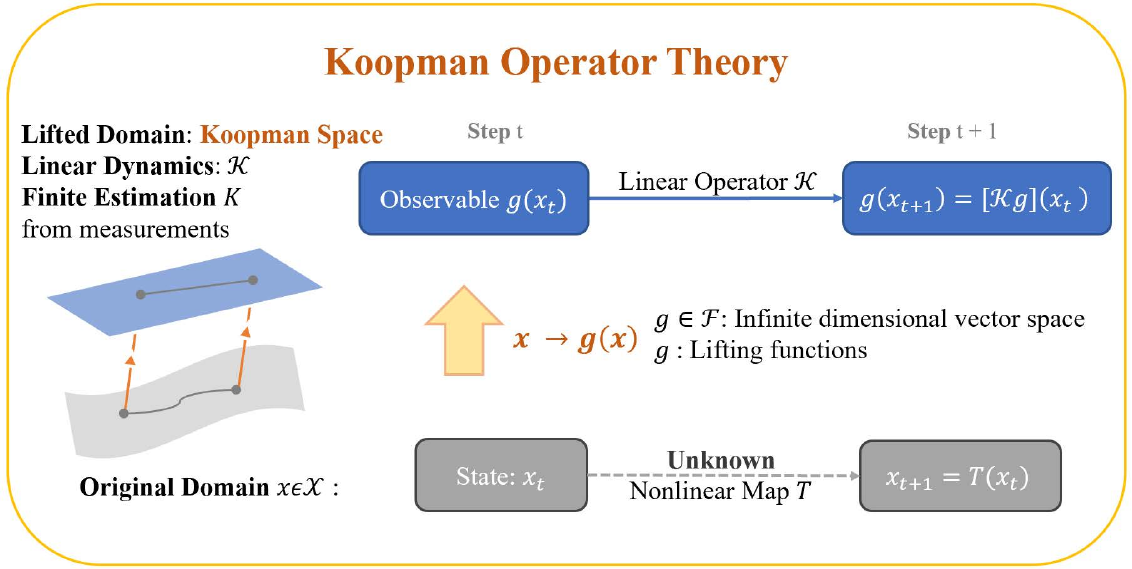}
    \vspace{-6pt}
    \caption{Conceptual illustration of Koopman Operator Theory. The nonlinear dynamics in the original state space $x_t \in \mathcal{X}$, governed by an unknown map $T$, are lifted via observable functions $g(x)$ into a higher-dimensional function space known as the Koopman space. In this lifted domain, the evolution of the system becomes linear and is governed by the Koopman operator $\mathcal{K}$, such that $g(x_{t+1}) = [\mathcal{K}g](x_t)$. Though $T$ and $\mathcal{K}$ act on different spaces, they evolve the same dynamics.}
    \label{fig:Koopman}
    \vspace{-9pt}
\end{figure}

Consider a discrete-time system\;\footnote{~Section~\ref{sec:advanced} elaborates on continuous-time dynamical systems.} expressed as 
\begin{equation}\label{eq:system}
    x_{t+1} = T(x_t)\;. 
\end{equation}
Let $x \in \Xc \subseteq \mathbb{R}^{N_{x}}$ be the state vector for the discrete-time dynamical system. The system's propagation rule is represented by the nonlinear function $T$ (Fig.~\ref{fig:Koopman}). 
Consider a vector space $\mathcal{F}$ of complex-valued functions (known as observables) with the system's state space as their domain. One can think of the observables $g \in \mathcal{F}$ as \textit{mapped} or \textit{lifted} states.
The evolution of the observables $g$ using the Koopman operator $\Kc: \Fc \to \Fc$ associated with system~\eqref{eq:system} is
\begin{align}\label{eq:Koopman-operator}
\mathcal{K} g= g \circ T, \quad \forall g \in \Fc\;,
\end{align}
where $\circ$ denotes function composition. The Koopman operator propagates the system one step forward in time according to  
\begin{align}\label{eq:temporal-evo}
\mathcal{K} g(x_t) = g \circ T(x_t) = g(T(x_t)) = g(x_{t+1})\;.
\end{align}
 
Through the process of lifting the original states $x$ using the observable functions $x \mapsto g(x)$, the system's dynamics can be redefined in a new space where the linear operator $\mathcal{K}$ describes the dynamics. 
In other words, although $T$ and $\mathcal{K}$ act on different spaces, they encapsulate the same dynamics. 
For example, given the current state $x$, one can propagate it to the subsequent time step and observe it through two pathways: either by employing $T$ to compute $T(x)$ and observe the resultant state (the bottom route in Fig.~\ref{fig:Koopman}) or by utilizing an observable function, applying $\mathcal{K}$ and evaluating it at $x$ (the top route). 
The concept of the ``equivalence" or ``substitution" offers several advantages. 
(1) It enables a global linear representation of the nonlinear dynamics $T$, thus enabling the application of techniques designed for linear systems. (2) It facilitates the discovery of the underlying dynamics by estimating the linear operator in real time, which eliminates the need for least-square regression of the nonlinear function that, in general, requires a large amount of data.


To ensure $\mathcal{K}$ is well defined,  $\mathcal{F}$ must be closed under composition with $T$. 
This condition might force $\Fc$ to be infinite-dimensional if one insists that $\Fc$ contains specific predefined functions (such as functions that return the full state values). 
To avoid dealing with infinite-dimensional spaces, a finite-dimensional representation of the Koopman operator can be employed instead. 
Thus, finite-dimensional subspaces of $\Fc$ that are invariant under the Koopman operator play a crucial role.\;\footnote{~$\Sc \subset \Fc$ is an invariant subspace under the Koopman operator if $\Kc g \in \Sc$ for all $g \in \Sc$.}  
Let $\Sc \subset \Fc$ be a finite-dimensional Koopman-invariant subspace and restrict the action of the Koopman operator to $\Sc$ as $\restr{\Kc}{\Sc}: \Sc \to \Sc$. Since $\Sc$ is finite-dimensional, given any basis for it, one can represent the operator $\restr{\Kc}{\Sc}$ by a matrix. Formally, let $\mathbf{\Psi}$ be a vector-valued function whose elements form a basis for $\Sc$, then there exists a matrix $K \in \cplx^{\dim(\Sc) \times \dim(\Sc)}$ such that 
\begin{align}\label{eq:Koopman-def}
\Kc \mathbf{\Psi} = \mathbf{\Psi} \circ T = K \mathbf{\Psi}\;,
\end{align}
where in the first term, the Koopman operator acts elements-wise on $\mathbf{\Psi}$. Combing~\eqref{eq:Koopman-def} and \eqref{eq:temporal-evo} yields a linear evolution of system trajectories, thus enabling the use of methods from linear systems theory for~\eqref{eq:system}, that is 
\begin{align}\label{eq:lifted-linear-system}
\mathbf{\Psi}(x_{t+1})  = K \mathbf{\Psi}(x_t)\;.
\end{align}
By defining $z_t := \mathbf{\Psi}(x_t)$,~\eqref{eq:lifted-linear-system} can be written as $z_{t+1} = K z_t$ which is in linear form. 
If the space $\Sc$ contains the state observables $g_i(x) = x^i$, where
$x^i$ is the $i$th element of state $x$, then one can choose the basis $\mathbf{\Psi}$ in a way that contains $g_i$'s as its elements.\;\footnote{~This is referred to as the full-state observability assumption~\cite{williams2015EDMD}. More generally, state observables can be written as a linear combination of elements of $\mathbf{\Psi}$, if they are contained in $\Sc$.} 
In this case, the lifted linear system~\eqref{eq:lifted-linear-system} captures the complete information of the original nonlinear system~\eqref{eq:system}. 
Moreover, the eigendecomposition of matrix $K$ allows one to identify the Koopman eigenfunctions and eigenvalues. 

\subsection{Data-driven Estimation and Prediction}
In practice, discrete measurement data can be used to construct a finite-dimensional matrix approximation of the infinite-dimensional Koopman operator. 
One of the most widely used methods for this approximation is the Extended Dynamic Mode Decomposition (EDMD)~\cite{williams2015EDMD}. 
Another approach, commonly adopted in robotics, particularly for soft robotic systems, is the Hankel View of Koopman (HVOK)~\cite{kamb2020HAVOK}. 
In the following, we will provide a detailed introduction to both methods.

\subsubsection{Extended Dynamic Mode Decomposition (EDMD)}\label{subsec:EDMD} Consider data matrices taken from system~\eqref{eq:system} as 
\begin{align*}
  X = [x_1,x_2, \ldots,x_M]\;,
  \\
  Y = [y_1,y_2,\ldots,y_M]\;,
\end{align*}
where $y_i = T(x_i)$ for all $i \in \until{M}$; we can also use the state history {$Y = [x_2,\dots,x_{M},x_{M+1}]$}. 


Define the action of the dictionary on a data matrix $X$ by $\Psib(X) = [\Psib(x_1), \Psib(x_2), \ldots, \Psib(x_M)]$. 
Then, EDMD can be equivalently formulated as a least Frobenius-norm problem
\begin{align}\label{eq:EDMD-optimization}
  \underset{K}{\text{minimize}} \| \Psib(Y) -  K \Psib(X)\|_F\;,
\end{align}
with the closed-form solution
\begin{align}\label{eq:EDMD-closed-form}
  \Kedmd = \Psib(Y)  \Psib(X)^\dagger\;.
\end{align}
The matrix $\Kedmd$ captures important information about the Koopman operator. Formally, given a function $f \in \Span(\Psib)$ denoted as $f(\cdot) = v_f^T \Psib(\cdot) = \sum_{i=1}^N (v_f)_i \psi_i(\cdot) $, the EDMD predictor for the function $\Kc f$ is defined as 
\begin{align}\label{eq:EDMD-function-predictor}
  \Pf_{\Kc f}^{\edmd} := v_f^T \Kedmd \Psib\;.
\end{align}
Note that $\Pf_{\Kc f} \in \Span(\Psib)$ even if $\Kc f \notin \Span(\Psib)$. An important special case of~\eqref{eq:EDMD-function-predictor} is for functions of the form $\phi(\cdot) = v_{\phi}^T \Psib(\cdot)$, where $v_{\phi}$ is a left eigenvector of $\Kedmd$ (that is, $v_{\phi}^T \Kedmd = \lambda_{\phi} v_{\phi}^T$), which leads to the approximated Koopman eigenfunction
\begin{align}\label{eq:EDMD-eigenfunction-predictor}
  \Pf_{\Kc \phi}^{\edmd} := v_{\phi}^T \Kedmd \Psib = \lambda_{\phi}
  v_{\phi}^T \Psib=  \lambda_{\phi} \phi\;. 
\end{align}

The approximation error of the predictors in~\eqref{eq:EDMD-function-predictor}-\eqref{eq:EDMD-eigenfunction-predictor} depends on the selected dictionary of functions~$\Psib$.  More precisely, the error depends on how close $\Span(\Psib)$ is to be invariant under the Koopman operator. To understand this, it is important to observe that $\Kedmd$ does not capture the Koopman operator itself but it encodes the projection of the operator's action on $\Span(\Psib)$, i.e. EDMD approximates the linear operator 
\begin{align}\label{eq:projection-Koopman}
  \Pc_{\Span(\Psib)} \Kc: \Fc \to \Fc\;,
\end{align}
where $\Pc_{\Span(\Psib)}: \Fc \to \Fc$ is the $L_2(\mu_X)$-orthogonal projection operator on $\Span(\Psib)$ and the $L_2$ inner product is calculated based on the empirical measure 
\begin{align}\label{eq:empirical-measure}
  \mu_X = \frac{1}{M} \sum_{i=1}^{M} \delta_{x_i}\;,
\end{align} 
with $\delta_{x_i}$ the Dirac measure at point~$x_i$. 

Since $\Span(\Psib)$ is always invariant under the operator in~\eqref{eq:projection-Koopman}, one can restrict the action of~\eqref{eq:projection-Koopman} to $\Span(\Psib)$ and represent the
restricted operator by a matrix (which, for exact data, coincides with $\Kedmd$)~\cite{SK-PK-CS:16,korda2018convergence}.  
This connection between EDMD and the Koopman operator leads to many important properties, including the convergence -- as the dictionary of basis functions grows -- of the
operator defined by EDMD matrix to the Koopman operator in operator topology, capturing the Koopman operator's eigenvalues, as well as weak convergence of eigenfunctions.  
We refer the reader to~\cite{korda2018convergence} for the detailed analysis and statements.

It is important to realize that none of the aforementioned convergence results imply that a larger finite-dimensional space is necessarily better for prediction. 
For example consider the linear system $x^+ = 0.5 x$ with two dictionaries $\Psib_1(x) = x$ and $\Psib_2(x) = [x, \sin(x)]$.  Despite $\Span(\Psib_1) \subsetneq \Span(\Psib_2)$, prediction on $\Span(\Psib_1)$ is exact (since it is invariant under the Koopman operator associated with the system). In contrast, the prediction on $\Span(\Psib_2)$ has large errors for some functions (see e.g.,~\cite{MH-JC:22-tac,MH-JC:21-tcns,MH-JC:23-auto} for methods that remove functions to prune subspaces and improve the prediction accuracy). 

In general, to observe the asymptotic convergence phenomena of EDMD to the Koopman operator described in~\cite{korda2018convergence}, the dimension of the dictionary and the number of data points must be \emph{sufficiently large}. Moreover, without a system's model, it is not possible to estimate a lower bound on the dictionary's dimension to achieve a predetermined level of accuracy. In addition, if the dictionary sequence is chosen based on a given basis for the space $\Fc$ (that is not chosen based on the system's knowledge), the dimension of the dictionary required to achieve a high accuracy level might be extremely large. Therefore, for practical applications, it is imperative to design or learn dictionaries based on information available from the system and/or data to achieve a reasonable accuracy on relatively low-dimensional subspaces. 

\begin{remark}\longthmtitle{Dynamic Mode Decomposition}
Dynamic Mode Decomposition (DMD) is a data-driven method originally proposed to extract coherent features of fluid flows~\cite{schmid2010dynamic}. Although developed before EDMD, the exact variant of DMD~\cite{Tu2014DMD} can be seen as a special case of EDMD where there is no lifting on the data. Alternatively, the EDMD dictionary can be set as the identity map to recover the exact DMD. 
\end{remark}

\subsubsection{Hankel View of Koopman Operator (HVOK)}
The Hankel View of Koopman (HVOK)~\cite{kamb2020HAVOK} is an alternative formulation of Koopman operator approximation that has shown effectiveness, particularly in systems with rich temporal dynamics such as soft robotics. Unlike EDMD, which directly learns a mapping between lifted state observations, HVOK leverages time-delayed embedding to implicitly capture the dynamics through the structure of the data.

Consider the dynamical system~\eqref{eq:system} and the data matrices $X$ and $Y$. 
While EDMD constructs a finite-dimensional Koopman approximation by applying a set of predefined basis functions \( \Psib \) to \( X \) and \( Y \), HVOK instead constructs `Hankel matrices' using stacked time-delayed snapshots of system measurements. The intuition is to embed temporal dynamics into a higher-dimensional space by capturing the evolution of observables over multiple time steps, enabling a more expressive representation of the underlying system.

Formally, the Hankel matrices are constructed as:

\[
H_X = \begin{bmatrix}
x_1  & \cdots & x_{m-d} \\
x_2& \cdots & x_{m-d+1} \\
\vdots  & \ddots & \vdots \\
x_d & \cdots & x_{m-1}
\end{bmatrix},
H_Y = \begin{bmatrix}
x_2  & \cdots & x_{m-d+1} \\
x_3  & \cdots & x_{m-d+2} \\
\vdots  & \ddots & \vdots \\
x_{d+1}  & \cdots & x_m
\end{bmatrix}.
\]

Then, similar to EDMD, HVOK seeks a linear operator \( K_{\text{HVOK}} \) such that \( H_Y \approx K_{\text{HVOK}}  H_X \). This operator captures the evolution of the system over time-delayed observables. The key difference is that HVOK does not require an explicit choice of basis functions—instead, it relies on the embedding power of delay coordinates, which can be seen as an implicit feature space.

\subsection{Koopman Approximation for Systems with Inputs}\label{subsec:inputs}
Extending the Koopman framework to handle control inputs is essential for its applicability in robotic control, planning, and estimation tasks. 
Rigorous operator-theoretic approaches to address this challenge include two formal methods discussed in Section~\ref{subsec:advancedInputs} and both approaches offer provable theoretical guarantees in the infinite-dimensional setting. 
In practice, however, directly working with infinite-dimensional spaces is computationally intractable. Therefore, a variety of finite-dimensional approximation strategies have been developed to embed input effects into the Koopman framework. These practical methods enable efficient and effective application in real robotic systems. In this section, we introduce several widely adopted approximation techniques for incorporating inputs into Koopman-based modeling. For a summary of representative works and the corresponding strategies they employ, please refer to Table~\ref{tab:literature review}.
\begin{enumerate}
    \item \textbf{Joint Lifting of States and Inputs}: A direct way to handle inputs is to treat the control input \( u \) as part of the system's extended state and define observables \( g(x, u) \) over this combined space. In this approach, the Koopman operator acts on functions of both the state and input, without assuming an independent evolution of \( u \). While intuitive and simple to implement, this method assumes knowledge of future inputs and does not generalize well when the input varies arbitrarily, since the input is not governed by a dynamical rule. Still, it has been effectively used in learning-based robotic systems with structured or repetitive input patterns.
    \item \textbf{Affine Input Form in the Lifted Space}: Another commonly used approach assumes that the evolution of observables can be modeled as affine in the input, i.e.
    \[
    g(x_{t+1}) \approx K g(x_t) + B u_t\;,
    \]
    where \( K \) is the finite-dimensional Koopman approximation for the autonomous part of the system, and \( B \) captures the linear influence of the control input. This method allows one to preserve the linear structure in the Koopman space while explicitly modeling controls. It aligns naturally with classical control design tools such as LQR and MPC. This formulation has been adopted in many robotic systems, especially when a linear control-friendly representation is desired. Interestingly, this approach is a particular case of the input-state separable model introduced in~\cite{MH-JC:24-auto}, which provides a solid theoretical backdrop to understand the errors associated with these  approximations.
    \item \textbf{Control-Coherent Koopman Operators}: Recently, the control-coherent approach~\cite{asada2024control} is proposed that focuses on `preserving consistency' in the Koopman embeddings under different control inputs. It aims to find an embedding space in which the evolution operator remains coherent, even as the control inputs vary. This method improves generalization to new control sequences and supports learning Koopman operators that are robust to input variability. It is particularly useful in manipulation tasks and underactuated systems where input variations are critical to task success.
    \end{enumerate}

Each of these methods provides a unique trade-off between model expressiveness, generalizability, and control compatibility. Their applicability often depends on the nature of the task, data availability, and system characteristics. Table~\ref{tab:literature review} provides a summary of representative works that employ these Koopman extensions in robotic systems, which readers may refer to for potential implementation strategies and inspiration.

\renewcommand{\arraystretch}{1.4}

\begin{table*}
    \caption{A summary of representative references and implementation details across various robotic systems.}
    \centering
    \begin{tabular}
    {c|c|c|c|c|c}
       \textbf{Robot Platforms}  & \multicolumn{2}{c|}{\textbf{Ways to Include Inputs}} & \textbf{Lifting Functions} & \textbf{Downstream Applications} & \textbf{Reference (s)} \\ \cline{2-3}
       & Inputs-affined & Jointly-lifted& & \\
\hline
\hline
       \multirow{5}{*}{\underline{\textbf{Manipulator}}}  &\checkmark & &\textcolor{NavyBlue}{$\circ$ Manually-selected} & \textcolor{red}{\ding{172}} MPC &\cite{hagane2023arm1} \\  \cline{4-6}
       
       & \checkmark& & \multirow{3}{*}{\textcolor{Plum}{$\blacksquare$ NN-based}} & \textcolor{red}{\ding{172}} MPC &\cite{zhang2023arm2}\\ \cdashline{5-6}
       
       & \checkmark& & & \textcolor{Lavender}{\ding{177}} Imitation Learning &\cite{bi2024arm7}\\ \cdashline{5-6}
       
       & \checkmark & &  & \textcolor{Lavender}{\ding{177}} Imitation Learning&\cite{han2023arm3}\\ \cline{4-6}
       
       & \checkmark& & \textcolor{BlueGreen}{\ding{228} DMD} & \textcolor{Lavender}{\ding{177}} Imitation Learning&\cite{chenkorol2024arm6}\\
\hline
\hline
       \multirow{6}{*}{\textbf{\underline{Wheeled Robots}}}& \checkmark & &\multirow{2}{*}{
       \textcolor{Blue}{\ding{170} Physics-informed}}& \textcolor{Yellow}{\ding{175}} NMPC &\cite{zhang2025ground1}\\ \cdashline{5-6}
       
       &\checkmark & & & \textcolor{red}{\ding{172}} MPC & \cite{joglekar2023ground2,rosenfelder2024ground5}\\ \cline{4-6}
       
       &\checkmark & &\multirow{4}{*}{\textcolor{NavyBlue}{$\circ$ Manually-selected}} & \textcolor{Gray}{\ding{178}} Robust Controller& \cite{ren2022wheeled}\\ \cdashline{5-6}
       
       &\checkmark & & & \textcolor{red}{\ding{172}} MPC&\cite{vsvec2023ground4} \\ \cdashline{5-6}
       
       & &\checkmark & & \textcolor{red}{\ding{172}} MPC& \cite{zhang2024ground3} \\ \cdashline{5-6}
       
       & &\checkmark & & \textcolor{Brown}{\ding{174}} State Observer &\cite{guan2025ground6} \\
\hline
\hline
       \multirow{3}{*}{\textbf{\underline{Legged Robots}}}&\checkmark & &\multirow{2}{*}{
      \textcolor{Plum}{$\blacksquare$ NN-based}} & \textcolor{red}{\ding{172}} MPC & \cite{li2024leg2,esfahani2024leg4}\\ \cdashline{5-6}
       
       &\checkmark & & & \textcolor{Brown}{\ding{174}} State Observer&\cite{khorshidi2024leg3}\\\cline{4-6}
       
       &\checkmark & & \textcolor{NavyBlue}{$\circ$ Manually-selected} & \textcolor{Orange}{\ding{176}} Modeling&\cite{krolicki2022quadruped}\\ 
\hline
\hline
       \multirow{13}{*}{\textbf{\underline{Soft Robots}}}  & & \checkmark &\multirow{4}{*}{\textcolor{NavyBlue}{$\circ$ Manually-selected}}
       & \textcolor{red}{\ding{172}} MPC & \cite{bruder2020koopman,thamo2022soft,peng2023soft,wang2023robust}\\ \cdashline{5-6}
       
       & &\checkmark & & \textcolor{Orange}{\ding{176}} Modeling& \cite{bruder2019nonlinear,han2025soft}\\ \cdashline{5-6}
       
       &\checkmark & & & \textcolor{Orange}{\ding{176}} Modeling& \cite{kamenar2020prediction}\\ \cdashline{5-6}
       
       &\checkmark & & & \textcolor{red}{\ding{172}} MPC & \cite{chen2022offset,wang2022improved,singh2023soft}\\ \cline{4-6}
       
       & &\checkmark & \multirow{3}{*}{\textcolor{Plum}{$\blacksquare$ NN-based}}& \textcolor{red}{\ding{172}} MPC &\cite{komeno2022soft}\\ \cdashline{5-6}
       
       &\checkmark & & & \textcolor{red}{\ding{172}} MPC & \cite{han2021desko,lilearning2019soft}\\ \cdashline{5-6}
       
       &\checkmark & & &\ding{180} Reinforcement Learning & \cite{ji2025soft}\\ \cline{4-6}
       
       & &\checkmark & \multirow{3}{*}{\textcolor{Blue}{\ding{170} Physics-informed}}& \textcolor{red}{\ding{172}} MPC&\cite{shi2022online}\\ \cdashline{5-6}
       
       & &\checkmark & & \textcolor{Orange}{\ding{176}} Modeling &\cite{shi2021ACDEDMD}\\ \cdashline{5-6}
       
       &\checkmark & &  & \textcolor{Orange}{\ding{176}} Modeling &\cite{castano2020control}\\ \cline{4-6}
       
       & &\checkmark & \textcolor{BlueGreen}{\ding{228} DMD}& \textcolor{Green}{\ding{173}} LQR & \cite{haggerty2023soft}\\ \cline{5-5} \cline{4-6}
       
       &\checkmark & & \multirow{2}{*}{\textcolor{CadetBlue}{\ding{73} HVOK}} & \textcolor{Green}{\ding{173}} LQR & \cite{haggerty2020modeling} \\ \cdashline{5-6}
       
       & &\checkmark & & \textcolor{Orange}{\ding{176}} Modeling& \cite{ristich2025soft,Vasudevan-RSS-19}\\
\hline
\hline
       \multirow{11}{*}{\textbf{\underline{Aerial Robots}}}& &\checkmark &
       \multirow{2}{*}{\textcolor{Blue}{\ding{170} Physics-informed}} 
        & \textcolor{Green}{\ding{173}} LQR & \cite{martini2024aerial4} \\ \cdashline{5-6}
        
        &\checkmark& & & \textcolor{red}{\ding{172}} MPC & \cite{narayanan2023aerial3}\\
       \cline{4-6}
        
        &\checkmark & & \multirow{3}{*}{\textcolor{NavyBlue}{$\circ$ Manually-selected}} 
        &   \textcolor{red}{\ding{172}} MPC & \cite{oh2024aerial5,folkestad2020episodic}\\ \cdashline{5-6}
        
         &\checkmark & & & \textcolor{Yellow}{\ding{175}} NMPC & \cite{folkestad2021koopmanNMPC} \\ \cdashline{5-6}
         
        & &\checkmark &   & \textcolor{Brown}{\ding{174}} State Observer & \cite{jia2023evolver}\\
         \cline{4-6}
        
        & &\checkmark & \multirow{2}{*}{\textcolor{CadetBlue}{\ding{73} HVOK}} 
           & \textcolor{red}{\ding{172}} MPC & \cite{louw2021aerial1} \\ \cdashline{5-6}
         
         &\checkmark & & & \textcolor{red}{\ding{172}} MPC & \cite{mamakoukas2022robust}\\
        \cline{4-6}
        
        &\checkmark & & \multirow{2}{*}{\textcolor{BlueGreen}{\ding{228} DMD}} 
           & \textcolor{red}{\ding{172}} MPC & \cite{jackson2023aerial2} \\ \cdashline{5-6}
         
         & &\checkmark & & \textcolor{Magenta}{\ding{179}} Adaptive Controller & \cite{shi2020CCTA}\\
        \cline{4-6}
        
        & &\checkmark & \multirow{2}{*}{{\textcolor{Plum}{$\blacksquare$ NN-based}}} 
         &  \textcolor{Brown}{\ding{174}} State Observer & \cite{huang2024SE}\\ \cdashline{5-6}
         
         &\checkmark & & &  \textcolor{Yellow}{\ding{175}} NMPC & \cite{folkestad2022koopnet} \\
\hline
\hline
        \multirow{4}{*}{\textbf{\underline{Underwater Robots}}}& &\checkmark & 
       \multirow{2}{*}{\textcolor{NavyBlue}{$\circ$ Manually-selected}} & \textcolor{Orange}{\ding{176}} Modeling& \cite{lin2024water2} \\ \cdashline{5-6}
       
       &\checkmark & &  & \textcolor{red}{\ding{172}} MPC & \cite{rodwell2023water1}\\ \cline{4-6} 
       
       &\checkmark& &\textcolor{Plum}{$\blacksquare$ NN-based}  &\textcolor{red}{\ding{172}} MPC & \cite{li2022learning}\\ \cline{4-6}
       
       &\checkmark & &\textcolor{Blue}{\ding{170} Physics-informed} &  \textcolor{Green}{\ding{173}} LQR & \cite{mamakoukas2021derivative}\\
\hline
\hline
        \multirow{2}{*}{\textbf{\underline{Rehabilitation Robots}}}&\checkmark & &
       \multirow{2}{*}{\textcolor{NavyBlue}{$\circ$ Manually-selected}}  & \textcolor{red}{\ding{172}} MPC & \cite{kalinowska2019limb}\\ \cdashline{5-6}
       
       & & \checkmark& & \textcolor{Yellow}{\ding{175}} NMPC & \cite{goyal2022wrist}\\
\hline
\hline
       \multirow{2}{*}{\textbf{\underline{Surgery Robots}}}& & \checkmark&
       \textcolor{NavyBlue}{$\circ$ Manually-selected} & \textcolor{Green}{\ding{173}} LQR & \cite{kong2023surgery}\\\cline{4-6}
       
       &\checkmark & &\textcolor{Blue}{\ding{170} Physics-informed}  & \textcolor{red}{\ding{172}} MPC & \cite{zhang2025needle}\\
\hline
\hline
    \end{tabular}
    \label{tab:literature review}
\end{table*}

\section{Koopman-Based Modeling and Its Application}\label{sec:model}
Unlike traditional model-based approaches that rely on explicit physical modeling or local linearization, Koopman-based methods aim to represent nonlinear dynamics as linear evolution in a higher-dimensional space of observables. 
This property makes it particularly attractive for robotics, where high-dimensional, nonlinear, and hybrid dynamics are common. 
To successfully apply Koopman theory in robotics, a typical pipeline involves three core stages: (1) collecting representative data from the robot's interaction with its environment, (2) constructing/selecting appropriate lifting functions to embed the original system into a space where linear evolution holds, and (3) applying the resulting model to `downstream' tasks such as controller design, state estimation through observers, and motion planning. In this section, we detail each of these components and discuss how they interconnect to form a complete Koopman-based modeling framework for robotics.  
An implementation overview of Koopman operator theory in robotic systems for control tasks is included in the supplementary tutorial (\url{https://shorturl.at/ouE59}). 
We also recommend investigating the impact of different dictionary selections and experimenting with various approaches to estimate the Koopman operator using the code provided by~\cite{pan2024pykoopman}. 
Additionally, designing and implementing different controllers can be explored using examples provided in~\cite{folkestad2021koopmanNMPC}, while the scripts used in~\cite{abraham2019active,folkestad2020episodic} offer insights into active data collection.

\subsection{Data Collection for Koopman Modeling}
As with all data-driven algorithms, the type and quantity of (training) data play a crucial role in model accuracy and control effectiveness. 
The quality of the training data significantly influences the ability to estimate an accurate representation of the underlying system, which in turn affects the performance of downstream tasks. 
Optimizing the data collection process in Koopman-based methods is therefore essential for effective learning and remains an active area of research~\cite{bakhtiaridoust2023data,mahmood2022optimizing}.

A common method to collect data is by random selection~\cite{chen2022offset}, i.e. generate measurements by propagating the system with random initial conditions and inputs. 
To enrich the diversity of the data, one should select initial conditions and inputs over the entire operating range of the robot~\cite{bruder2021advantages,bakhtiaridoust2023data}. 
An enclosed area might be required for safety~\cite{guo2021koopman}. 
The random collection approaches are more widely used for soft robots~\cite{castano2020control,bruder2020data,wang2022improved} as soft parts do less harm to their surroundings if an aggressive command is selected by chance. 

To mitigate the safety concerns associated with random action selection, another typical data collection method is to start with a baseline controller—possibly open-loop and naive—and iteratively refine it over time.
For instance, Folkestad et al.~\cite{folkestad2020episodic} employ a scheme in which the controller from the previous episode is used to generate data for the current one. 
This iterative process enables safer data collection while also improving the informativeness of the observations. 
Another approach to improve data efficiency is to explicitly consider the value of information during data collection. 
This can be achieved by introducing constraints on information richness in the optimization process. Such strategies are closely related to active learning paradigms~\cite{abraham2019active}, which we will discuss in more detail in Section~\ref{subsubsec:activeLearning}.

\subsection{Lifting Functions Selection for Koopman Embedding}\label{subsec:modelEst}
Once the dataset has been collected, the next critical step 
is the design of an appropriate dictionary of lifting functions, which are used to map the original system states (and possibly inputs) into a higher-dimensional space where the dynamics can be approximated linearly. These lifting functions form the foundation for constructing the finite-dimensional approximation of the Koopman operator.

Broadly speaking, existing strategies for constructing lifting functions can be categorized into three main directions.
\begin{enumerate}
    \item \textbf{Manually-selected Basis Functions}: This approach involves choosing a set of basis functions informed by domain knowledge or trial-and-error. Examples include using spectral elements that lead to block-diagonal observation matrices~\cite{williams2015EDMD}, Hermite polynomials for normally distributed data, or radial basis functions (RBFs) for systems with complex or spatially structured dynamics. While effective in some cases, empirical design is often labor-intensive and may not generalize well to different systems or tasks.
    \item \textbf{Physics-informed Lifting Functions}: Robots often exhibit structured properties such as kinematic constraints, degrees of freedom, and geometric configuration spaces that can be leveraged in lifting function design, even in the absence of full dynamic models. 
    For example, Shi et al.~\cite{shi2021ACDEDMD} propose an approach that incorporates physical insight, such as configuration symmetries or workspace constraints, into the construction of the observable space, improving both the interpretability and robustness of the resulting Koopman model. 
    Another example is to synthesize basis functions using higher-order time state derivatives~\cite{mamakoukas2021derivative}, which shows how including derivative terms in the observable space can enrich the expressiveness of the Koopman approximation. More representative works are listed in Table~\ref{tab:literature review}.
    \item \textbf{Neural Network-based Lifting Functions}: Learning lifting functions directly from data using neural networks~\cite{li2017ML} is also possible. 
    Such approaches often fall under the umbrella of Deep Koopman~\cite{shi2022deep} or Autoencoder-Koopman frameworks~\cite{asada2023global}, where neural networks are trained to learn an embedding that captures the latent linear dynamics.
    While NN-based lifting functions offer high flexibility and expressiveness, they also introduce challenges such as interpretability, generalization to out-of-distribution inputs, and the risk of overfitting.
\end{enumerate}

The selection of lifting functions plays a critical role in the effectiveness of Koopman-based modeling, as it directly influences the accuracy of the operator approximation. 
The choice among the three aforementioned categories of methods should be guided by the characteristics of the system and the specific task at hand. 
As summarized in Table~\ref{tab:literature review}, different robotic platforms tend to favor different approaches. 
For example, NN-based methods are widely adopted in manipulation and legged locomotion due to the high complexity and nonlinearity of their dynamics. 
In contrast, manually designed functions are more commonly used in wheeled robot systems, where the dynamics are comparatively simpler and better understood. 
In the case of aerial and soft robots, all types of lifting functions have been explored, but it is noteworthy that the HVOK framework is particularly favored due to its ability to capture strong environmental disturbances (as seen in aerial systems) or slow response characteristics (as in soft robotics). 
While we present these widely used and practical methods, it is important to note that most lack rigorous convergence analysis—that is, they do not examine whether the constructed observables span a Koopman-invariant subspace or yield an accurate approximation of the true Koopman operator. 
To address this gap, we provide a more in-depth theoretical discussion on the construction of lifting functions in Section~\ref{subsec:advancedLifting}.

\subsection{Application in Robot Controller Design}~\label{subsec:control}
Once a finite-dimensional approximation of the Koopman operator is obtained, it can be seamlessly integrated into control design, enabling efficient handling of nonlinear robotic dynamics. Powerful model-based control methods—such as Linear Quadratic Regulation (LQR), Model Predictive Control (MPC), and Nonlinear MPC (NMPC)—within a data-driven setting can be applied. Moreover, the data-centric nature of Koopman-based modeling naturally supports active learning strategies, in which the control policy not only drives the system toward its objective but also enhances data efficiency by improving model accuracy through exploration.

In this review, we focus on these two foundational applications of Koopman theory in robotics. Beyond these, a growing body of work explores other promising directions, including adaptive control~\cite{shi2020CCTA}, robust control~\cite{ren2022wheeled}, and reinforcement learning, where Koopman models are used to either approximate environment dynamics~\cite{ji2025soft} or support the design of critic networks~\cite{rozwood2023koopmanRL}. Due to space limitations, we do not expand on these additional topics here, but representative examples can be found in Table~\ref{tab:literature review}.
\subsubsection{Model-based Controller}
Building upon the foundational concept of embedding system dynamics into a higher-dimensional function space via the Koopman operator, the integration of Koopman theory into model-based control frameworks marked a significant turning point. 
This idea was first introduced in~\cite{abraham2017RSS}, laying the groundwork for a new class of data-driven controllers that leverage linear representations of nonlinear dynamics. 
Subsequently, this framework evolved into the development of Koopman Model Predictive Control (Koopman MPC)~\cite{korda2018KoopmanMPC} and its nonlinear extension, Koopman Nonlinear Model Predictive Control (Koopman NMPC)~\cite{folkestad2021koopmanNMPC}. 
These controllers exploit Koopman-based models—derived from either real-time measurements or offline datasets collected from physical systems—to achieve effective control even in the absence of complete system knowledge.

At the core of these approaches lies the optimal control formulation, where control actions are computed by solving a constrained optimization problem over a finite prediction horizon, that is
\begin{align}\label{eq:mpc}
    &\underset{\{{\bm{z}}_i\}_{i=0}^{N_h} , \{{\bm{u}}_i\}_{i=0}^{N_h}}{\text{minimize}} 
    && J( \{{\bm{z}}_i\}_{i=0}^{N_h} , \{{\bm{u}}_i\}_{i=0}^{N_h}) 
    \notag
    \\
    &\hspace{10pt}\text{subject to}&&{\bm{z}}_{i+1} = F_\mathcal{K} \left( {\bm{z}}_i , {\bm{u}}_i \right) 
    \notag
    \\
    & &&{\bm{z}}_0 = \mathbf{\Psi}(\bm{x}_t)\;,
\end{align}
where $N_h$ is the number of time steps in the controller horizon, $J$ is a quadratic cost function, $\mathbf{\Psi}$ is the lifting dictionary, $F_\mathcal{K}$ is the Koopman-based system model, and $\bm{z}_i$ and $\bm{u}_i$ are the lifted system state and input at the $i^\text{th}$ time-step in the horizon, respectively. The problem’s objective function penalizes deviations from a desired trajectory, and its constraints enforce consistency with a Koopman system model. 
With a linear Koopman model realization, the optimization~\eqref{eq:mpc} becomes a quadratic program (QP) 
\begin{align}\label{eq:k-mpc}
    & \underset{\{{\bm{z}}_i\}_{i=0}^{N_h} , \{{\bm{u}}_i\}_{i=0}^{N_h}}{\text{minimize}} \hspace{-6pt}
    \notag
    &&\sum_{i=0}^{N_h} \bm{z}_i^\top G_i \bm{z}_i + \bm{u}_i^\top H_i \bm{u}_i + \bm{g}_i^\top \bm{z}_i + \bm{h}_i^\top \bm{u}_i 
    \notag\\
    &\hspace{10pt}\text{subject to}& &{\bm{z}}_{i+1} = K {\bm{z}}_i + B {\bm{u}}_i 
    \notag\\
    && &{\bm{z}}_0 = \mathbf{\Psi}(\bm{x}_t)\;.
\end{align}

A major advantage of a linear Koopman model realization is that it makes the formulation to be convex. 
This implies it has a unique globally optimal solution that can be efficiently computed without initialization even for high-dimensional models~\cite{boyd2004convex, paulson2014fast,stellato2018osqp}, making Koopman MPC well-suited for real-time feedback control applications. 
Nonlinear/bilinear Koopman model realizations render \eqref{eq:mpc} non-convex, which are less efficient to solve and may only yield locally optimal solutions~\cite{polak2012optimization}. 
Sometimes a nonlinear Koopman realization generates more accurate predictions, in which case such a trade-off is warranted. 
Bilinear Koopman model realizations have been explored recently (e.g.,~\cite{zhang2025ground1}) to combine some of the advantages of linear and nonlinear models~\cite{bruder2021advantages}.


\subsubsection{Active Learning} \label{subsubsec:activeLearning}
The linear structure of the Koopman operator yields several advantages for learning the dynamics of a robotic system. 
The closed-form solution of the linear least-squares model fitting problem~\eqref{eq:EDMD-optimization}, and \eqref{eq:EDMD-closed-form} to approximate the Koopman operator can be leveraged to formulate active-learning controllers~\cite{abraham2019active}. 
Specifically, assume that the Koopman dynamics from the dictionary of observables $\Psib(x_{t+1}) = K\Psib(x_t) + B\bm{u}_k$ form the mean of a normally distributed state in the lifted space $p(\bm{z}_{t+1} | K, \bm{z}_k )$. 
Then, one can form an approximation of the \emph{Fisher-Information Matrix}
\begin{equation}
\hspace{-3pt}
    \mathbf{I} = \mathbb{E} \left[ \frac{\partial}{\partial K} \log p(\bm{z}_{t+1} | K, \bm{z}_k ) \frac{\partial}{\partial K} \log p(\bm{z}_{t+1} | K, \bm{z}_k )^\top \right]\hspace{-1pt}.\hspace{-3pt}
\end{equation}
For normally distributed systems with zero-mean and variance $\Sigma$, this attains the closed-form expression
\begin{equation}
    \mathbf{I} = \frac{\partial \bm{z}_{t+1}}{\partial K}^\top \Sigma^{-1}\frac{\partial \bm{z}_{t+1}}{\partial K} \le \text{Var}[K^\star]^{-1}\;,
\end{equation}
where $\bm{z}_{t+1} = K \bm{z}_t  + B \bm{u}_t$, and $\text{Var}[K^\star]^{-1}$ is the posterior variance of the approximate Koopman operator. 
The Fisher-information matrix lower-bounds the posterior uncertainty in estimation, defined as the Cram\'er-Rao bound~\cite{cramer1946contribution, radhakrishna1945information}. 
Further, it is differentiable and actionable, i.e. one can optimize controllers that directly maximize the Fisher information, thus improving the best-case posterior variance on the Koopman operator, given the current operator. 

Such controllers can be obtained via optimization using the optimality conditions on the Fisher-information~\cite{radhakrishna1945information}, i.e. 
\begin{align}\label{eq:active_learning_opt}
    & \underset{\{\bm{u}_i\}_{i=0}^{N_h} }{\text{minimize}}&&\sum_{i=0}^{N_h-1}\mathfrak{I}(\bm{z}_i, {^t}K) + \bm{u}_i^\top R \bm{u}_t 
    \nonumber\\
    &\text{subject to } && \bm{z}_{i+1} = {^t}K \bm{z}_i + {^t}B \bm{u}_i \nonumber \\ 
    & && \bm{z}_0 = \Psib(x_t) 
\end{align}
where $R \succ 0$ is a positive definite matrix, $\mathfrak{I}(\bm{z}_i, {^t}K) $ is an optimality condition (e.g., D-, or T- Optimality) that reduces the matrix into a scalar value, $N_h$ is the time-horizon, and the superscript $^t$ indicates the current estimate of the Koopman operator given state-control data collected in the past. Note that the above optimization problem is done in a receding-horizon manner to account for changes in the Koopman operator. Controllers that are formulated from the optimization~\eqref{eq:active_learning_opt} effectively lower the overall posterior variance of the model yielding an effective Koopman operator model with a few data points. 
The resulting model can be used across different tasks such as for an aerial robot to quickly recover from an unstable tumble or a legged robot to learn complex interaction models with granular media (see~\cite{abraham2019active}). 
    
The simple structure of the Koopman operator has advanced learning in other directions. 
In particular, the use of deep models to approximate the function observables has made significant strides in expanding the use of methods based on the Koopman operator~\cite{abraham2019active, yeung2019learning, xiao2022deep}. 
Seep function observables can further be integrated into an active-learning problem with some success~\cite{abraham2019active}. 
While the added complexity in the observables provides more flexibility in the modeling range of the Koopman operators, it does reduce the effectiveness of active learning. This is a result of more data being required to effectively learn the nonlinear observables. 
When compared to deep neural network models, the Koopman-based linear model still has a significant advantage in data-efficiency and control through active learning~\cite{abraham2019active}. 

\subsection{Application in Robot State Estimation}

Building upon the success of Koopman operator theory in model-based control, recent research has extended its utility to the domain of state estimation, where the focus shifts from designing control actions to accurately recovering system states under uncertainty, disturbance, or partial observability. 
Existing research can be categorized into (1) robust state estimation across system variability, (2) disturbance estimation and rejection, and (3) efficient state inference in nonlinear or high-dimensional dynamics.

In the first direction, efforts focus on achieving robust estimation despite model uncertainties or population variability. For example,~\cite{dahdah2024SE} proposes a robust nonlinear observer synthesis method for a population of systems modeled using Koopman operators. By exploiting the linear representation in the lifted space, uncertainty due to manufacturing variations is quantified in the frequency domain, allowing the use of mixed \( \mathcal{H}_2/\mathcal{H}_\infty \) robust control techniques to design stable observers across dozens of motor drives.

The second direction deals with disturbance estimation, aiming for both fast reaction and steady performance. The EVOLVER framework in~\cite{jia2023evolver} draws inspiration from biological systems to handle internal and external disturbances. It utilizes Koopman-based latent structure modeling within an evolutionary disturbance observer, enabling rapid transient reactions and high-precision steady-state estimation, with convergence guarantees under optimal conditions.

The third direction focuses on data-driven and computationally efficient state estimation for nonlinear systems. 
To this end,~\cite{jiang2022SE} presents a data-driven Kalman filter for unknown nonlinear robotic systems by leveraging a sparse kernel-based Koopman operator, which transforms the system into a linear dynamic model in kernel space, allowing standard linear Kalman filtering to be applied. 
KoopSE~\cite{guo2021koopman} presents a batch state estimation method for control-affine systems that avoids linearization and manual feature selection. 
It lifts dynamics into a Reproducing Kernel Hilbert Space, making the system bilinear and allowing standard linear estimation techniques to operate efficiently. 
Random Fourier Features are employed to retain generalization while maintaining low inference cost. 
Similarly,~\cite{huang2024SE} introduces K-ESKF, a Koopman-enhanced error-state Kalman filter for agile quadrotor pose estimation. 
A deep neural network learns Koopman observables to convert full-state nonlinear dynamics into a bilinear control system, improving propagation accuracy in the ESKF framework. 

\subsection{Application in Robot Planning and Localization}
Koopman operators in planning often involve estimating a dynamics model (at times via data-driven lifting techniques) and incorporating it into an MPC framework~\cite{van2021deepkoco}. 
This usage parallels the controller design described previously, leveraging Koopman's linear representation to handle nonlinear planning tasks in a computationally efficient way~\cite{liang2025safe}. Beyond this general usage, recent research has extended Koopman theory to tackle more complex problems. 
One important direction focuses on improving data-driven localization and mapping. 
For instance, Koopman linearization has been used to reformulate the batch SLAM problem by lifting the system dynamics into a higher-dimensional space where both the process and measurement models become bilinear~\cite{guo2024data}. 
This formulation allows simultaneous estimation of robot trajectories and landmarks through a constrained optimization process, offering an elegant way to maintain consistency on the learned Koopman manifold during inference.

Another key challenge lies in navigation through complex, unstructured environments. 
To address this, Koopman operators have been used to model terrain traversability by incorporating features such as elevation, slope, and roughness. 
By lifting the navigation problem into density space using the linear dual of the Koopman operator, the problem becomes convex and more tractable for optimization~\cite{moyalan2023convex}. 

Finally, in scenarios with high parametric uncertainty, Koopman operators have been applied to develop uncertainty-aware motion planners~\cite{gutow2020koopman}. 
These planners rely on the Koopman framework to compute expected values and chance constraints for probabilistic collision avoidance. 
By integrating motion primitives with Koopman-based expected cost evaluations, this line of work enables maneuver-based planning strategies that can effectively balance risk and performance.

\subsection{Studies on Robustness and Stability}
While Koopman-based modeling offer a promising framework through linearization in lifted spaces, their performance can be significantly affected by inaccuracies in the learned models. 
These errors may arise from noisy measurements, limited observability, data scarcity, or the inability of finite-dimensional approximations to fully capture complex real-world nonlinear dynamics~\cite{zhang2019DMDaccuracy}. 
As a result, ensuring robustness to such model uncertainties is key, particularly for applications requiring safety, stability, and real-time performance.

The first source of uncertainty stems from the data collection process. When learning system dynamics from data, measurement noise can significantly degrade model quality. In stochastic environments, both the quantity and quality of data affect the accuracy of the learned Koopman operator and, consequently, the effectiveness of the controller. To quantify the influence of noisy data, Shi et al.~\cite{shi2020CCTA, shi2021noise} derive both loose and tight bounds on prediction errors for models obtained via DMD and EDMD. These error bounds can then be integrated into the control design to explicitly account for uncertainty during planning and execution.

At the model level, various strategies have been proposed to enhance robustness. Chen et al.\cite{chen2022offset} introduce a Kalman filter-enhanced Koopman framework, where an augmented model includes a disturbance term, allowing for joint estimation of lifted observables and unknown disturbances. This improves resilience to unmodeled dynamics and external noise. In a different approach, Han et al.\cite{han2021desko} develop the deep stochastic Koopman operator by learning a distribution over observables using a probabilistic neural network. The Koopman matrices then propagate this distribution, and both components are trained jointly to handle stochastic systems effectively. Further, Mamakoukas et al.~\cite{mamakoukas2023learning} propose computing the nearest stable Koopman matrix to reduce reconstruction error while promoting inherent stability in the learned model.

Beyond improving model fidelity, robust control design can further mitigate the effects of model mismatch. Wang et al.\cite{wang2022KTMPC} address modeling errors by proposing a constraint-tightening strategy within a tracking MPC framework. This guarantees recursive feasibility and input-to-state stability under bounded uncertainty. Kochdumper et al.\cite{kochdumper2022conformant} take a complementary route by constructing a conformant Koopman model that enforces both trace and reach-set consistency with the real system. Additionally, Mamakoukas et al.~\cite{mamakoukas2022robust} estimate the Lipschitz constants of model error with respect to both states and controls. These constants are then used to bound the prediction error over the control horizon, enabling the formulation of a robust MPC controller that adapts to uncertainty during online execution.


\section{Implementation of Koopman Methods in Different Robotic Systems}\label{sec:Robots}
The theoretical foundations provide a general framework for lifting nonlinear dynamics into linear representations; the practical implementation of these methods must be tailored to the unique characteristics and constraints of different robotic platforms. In this section, we survey how Koopman operator theory has been instantiated in diverse robotic systems, which aims to provide practical insights into the versatility and applicability of Koopman methods across the robotics domain. A summary of representative implementations across various robotic systems is provided in Table~\ref{tab:literature review}, highlighting the specific system types, learning objectives, Koopman formulations, and control strategies adopted in each case.

\subsection{Robotic Manipulation}\label{subsec:manipulator}
(Rigid) Robot manipulators are used extensively owing to their feasibility, stability, and safety. 
Recent work has explored a variety of ways to apply Koopman theory to manipulation. 
One prominent category focuses on modeling and predictive control for robotic arms. 
For instance,~\cite{hagane2023arm1} presents a Koopman-linearized model of a robotic manipulator that is integrated with a generalized predictive control (GPC) framework. 
This combination allows for effective planning and execution in systems that would otherwise require nonlinear control techniques.
The overall approach is further extended in~\cite{zhang2023arm2}, by developing a structured deep Koopman model that leverages deep neural networks with Lipschitz constraints to enhance the accuracy and interpretability of the lifted model while minimizing dimensionality. 
Similarly,~\cite{sah2024arm4} introduces a Koopman-Zeroing Neural Network architecture, combining an autoencoder-based Koopman model with a Cartesian-space feedforward network to achieve real-time control of redundant manipulators, even under input constraints.

Another category leverages Koopman operators to improve reinforcement and imitation learning in manipulation. In~\cite{sinha2022humanrobot}, human-demonstrated trajectories are encoded into a human intent model via Koopman lifting, which is then used to shape reward functions for reinforcement learning agents, enabling safe and task-constrained human–robot interaction.
A deep Koopman framework to learn a compact latent representation of system dynamics from observation-only data is developed in~\cite{bi2024arm7}. 
A linear decoder then maps these latent representations to real-world actions, drastically reducing the amount of action-labeled data required for imitation learning. 

A third category applies Koopman-based modeling to object-centric and dexterous manipulation. 
The KOROL framework~\cite{chenkorol2024arm6} allows for to extraction of visual object features and applies Koopman rollouts to predict future trajectories in feature space. 
These predicted trajectories are then used by an inverse dynamics controller to generate precise manipulation actions, offering interpretability and robustness in vision-based manipulation. 
Notably,~\cite{han2023arm3} marks an important shift toward dexterous hand manipulation, presenting a Koopman-based imitation learning framework that jointly models the dynamics of both the robotic hand and the manipulated object. 
The joint Koopman lifting captures the entangled dynamics between hand and object, supporting more accurate and generalizable skill learning.

\subsection{Ground Robots}\label{subsec:ground}
As robots continue to move from operation in structured environments to more unstructured and natural terrains, effective locomotion, both in wheeled and legged robots, has become a critical capability. 
Wheeled robots, while simpler in mechanics compared to their legged counterparts, are still governed by complex nonlinear dynamics due to continuous contact with the environment. 
These effects are pronounced when traversing deformable or uneven terrains, where modeling uncertainties and disturbances become prominent~\cite{borges2022survey}. 
Koopman operator-based approaches offer a data-driven alternative, enabling model-based controllers to be applied even in the absence of first-principles models. 
For example, Koopman-based MPC has been successfully deployed for wheeled robot navigation over complex terrain~\cite{wang2022KTMPC}, while enhancements like virtual control inputs~\cite{ren2022wheeled} can improve handling of rotational and coordinate-transformed dynamics in mobile platforms.

Legged robots present even greater challenges for modeling and control~\cite{carpentier2021legged}. 
Their dynamics are inherently hybrid, involving discrete contact switching events, underactuated behaviors, and high degrees of freedom. 
Accurate and generalizable modeling of these systems remains an open problem. 
In this context, Koopman operator theory has emerged as a promising tool to extract linear approximations of these nonlinear systems in a globally valid manner. 
An initial attempt at modeling legged robots with the Koopman operator is done with the quadruped leg dynamics on deformable terrains in~\cite{krolicki2022quadruped}. 
An experimental framework has been proposed to obtain a data-driven Koopman-based model of a quadruped's leg dynamics as a switched dynamical system. 
Two recent studies!\cite{ordonez2024leg1,li2024leg2} focus on modeling the full-body or local leg dynamics of legged robots using Koopman embeddings. 
By capturing global linearizations of the hybrid motion dynamics, they enable prediction and analysis across different gaits and terrains. 
In particular,~\cite{li2024leg2} extends this by integrating the learned Koopman model into an MPC framework, demonstrating effective closed-loop control of the legged robot, while also addressing domain shift through an incremental learning scheme that progressively refines the Koopman embedding and latent space as new data is collected. 
Other works, such as~\cite{khorshidi2024leg3} and~\cite{esfahani2024leg4}, take on a different approach, focusing on how to leverage Koopman operator theory at a higher abstraction level. 
Rather than directly modeling low-level dynamics, these studies employ Koopman embeddings to formulate estimation and planning strategies for high-level states of the robot. 
Although current developments in this area are still in their early stages, with most evaluations conducted primarily in simulation, research activity is rapidly growing, and Koopman-based approaches are poised to become a powerful tool for modeling and control in legged locomotion.

Distinctively yet crucially, the Koopman operator has also found its way into the realm of autonomous driving~\cite{manzoor2023AUVreview}. 
A key challenge is how to identify a global vehicle dynamics model considering the inherent complexity of the different subsystems and the induced nonlinearities and uncertainties. 
Considerable efforts have been directed toward utilizing the Koopman operator theory for estimating vehicle models~\cite{cibulka2019model}. 
The resulting approximated models can then be integrated into controller design, such as MPC~\cite{vsvec2021MPC,sheng2022MPC2,yu2022MPC3}. 
In some cases, like in operation over deformable terrains with significant height variations and the presence of bumps~\cite{buzhardt2022terrain}, it has been shown possible to obtain a linear representation of both vehicle and terrain interaction dynamics using Koopman estimation. 
Further advancements have considered bilinear model formations for autonomous vehicles~\cite{yu2022autonomous} and their combination with deep neural networks~\cite{xiao2022deep}. 
The integration of driver-in-the-loop dynamics has been explored in~\cite{guo2023driver}, reflecting a holistic approach to autonomous driving. 
Additionally,~\cite{chakraborty2023attention} introduces the embedding of the stochastic Koopman operator structure within an attention-based framework. 
This is leveraged for abnormal behavior detection in autonomous driving systems, thereby enhancing the safety of vehicle controllers. 
Readers can refer to the comprehensive and detailed survey on the application of Koopman operator theory in autonomous vehicles presented in~\cite{manzoor2023survey}.

\subsection{Soft and Continuum Robots}\label{subsec:soft}
Soft robots have emerged enjoy high compliance, adaptability to complex environments, and inherent safety in interacting with humans and surroundings. 
However, their highly nonlinear and complex dynamics present substantial challenges for traditional modeling and control techniques. 
Koopman operator theory offers a powerful data-driven alternative, enabling global linear representations of these otherwise intractable dynamics. 
Moreover, the physical safety of soft robots makes it feasible to collect large volumes of rich data under diverse control inputs without the risk of damage—an advantage not easily attainable with rigid-body robots. 

These characteristics make soft robots one of the most promising application domains for Koopman operator theory. 
Notably, research activity in this direction has surged in the past two years, with growing efforts focused on both modeling~\cite{bruder2019nonlinear,kamenar2020prediction,han2025soft,ristich2025soft} and closed-loop control~\cite{shi2023review}. 
Koopman theory has been successfully applied to a variety of soft robotic systems, enabling real-time or near-real-time control performance through globally linear representations of their otherwise nonlinear dynamics~\cite{bruder2020koopman}. 
As summarized in Table~\ref{tab:literature review}, various methods have been explored for constructing lifting functions to embed soft robot dynamics into higher-dimensional linear systems. 
These include polynomial basis functions~\cite{zhou2025soft}, time-delay embeddings~\cite{haggerty2020modeling,Vasudevan-RSS-19}, and neural network-based approaches~\cite{komeno2022soft,han2021desko}. 
In terms of control strategies, MPC and LQR are most widely adopted due to their compatibility with the Koopman-linearized models (for example, see~\cite{shi2022online} and~\cite{haggerty2023soft}, respectively). 
Beyond standard control pipelines, an intriguing recent application involves using a Koopman-based model as a surrogate environment for training reinforcement learning (RL) policies~\cite{ji2025soft}. 
This approach bypasses the need for difficult-to-construct physical or simulated soft robot environments, leveraging the Koopman model to accelerate policy learning in complex morphologies.

\subsection{Aerial Robots}\label{subsec:aerial}
The inherent nonlinearity of aerial robots presents significant challenges for achieving precise control. 
Key challenges arise when there is interaction with the environment that is governed by hard-to-model aerodynamic effects~\cite{karydis2017energetics}, such as variable wind gusts and ground effects when landing~\cite{shi2019neural} or flying at a low altitude~\cite{karydis2017uncertainty,kan2019analysis}, which can corrupt the nominal model and lead to unexpected performance. 


Data-driven Koopman-based approaches have attempted to mitigate the effect of such uncertain aerodynamic interactions 
by extracting and adapting online system models used to linearly control the aerial robot. 
This overall idea can be achieved in different ways. 
For example,~\cite{folkestad2020episodic} uses episodic learning to estimate the Koopman eigenfunction pairs and obtain the resulting control inputs in real time to handle the ground effect when a multirotor is landing. 
A sequence of Koopman eigenfunctions is iteratively learned from data. 
Accordingly, nonlinear control signals are improved from the nominal control law. 
Another implementation of the Koopman operator theory in aerial robots is to jointly learn a function dictionary and the lifted Koopman bilinear model to achieve quadrotor trajectory tracking at a low altitude~\cite{folkestad2022koopnet}. 
A neural network is combined with the Koopman operator to update both the lifted states and inputs of the robot with online measurements. 
In a different effort, a hierarchical structure is proposed that refines the reference signal sent to the high-rate, pre-tuned low-level controller of the aerial robot to deal with uncertainty~\cite{shi2020CCTA}. 
A model of the reference and the actual output of the disturbed robot is learned via Koopman operator theory and is utilized in the outer controller to decrease the effect of environmental disturbance in real time. 



\subsection{Other Types of Robot Platforms}\label{subsec:other}

Beyond the previously discussed categories, Koopman operator theory is beginning to make inroads into a broader range of robotic platforms, offering new possibilities for modeling and control in complex, nonlinear environments. 
For instance, in the domain of autonomous robotic excavation, the interaction between the excavator bucket and the surrounding soil is governed by highly nonlinear dynamics. 
These interactions have been effectively approximated using Koopman-based linear models, enabling simplified representations that support real-time control and planning~\cite{selby2021excavation,sotiropoulos2021excavator}. 
Similarly, building on foundational work that applied the Koopman operator to fluid flow analysis~\cite{mezic2013analysis}, researchers have begun to explore its potential in underwater robotics, where vehicles must adapt to unsteady and dynamic flow fields~\cite{mamakoukas2021derivative}. 
In such contexts, Koopman-based formulations enable linear approximations of otherwise nonlinear hydrodynamic interactions, which can be directly integrated into model-based controllers as linear constraints~\cite{li2022learning,bevanda2022towards}. 

Applications of the Koopman framework are also emerging in the area of rehabilitation robotics and assistive devices~\cite{zhou2025learning,goyal2022wrist,kalinowska2019limb}. These systems often experience hybrid dynamics due to discrete transitions—such as changes between contact and non-contact modes—which traditionally complicate modeling and control. Koopman-based approaches offer a promising solution by transforming these heterogeneous, mode-switching systems into a unified global linear model~\cite{asada2023global}, thus facilitating the design of effective controllers for dynamically shifting conditions. Other notable examples include Koopman-based model extraction in snake-like locomotion~\cite{zhu2022snake}, the coordination of smarticle ensembles~\cite{savoie2019smarticle}, and even applications in the control of surgical robots~\cite{kong2023surgery}.

\subsection{Multi-agent Systems}
In addition to its applications in single-agent systems, Koopman operator theory has proven valuable in the domain of multi-agent systems. 
A comprehensive introduction and tutorial on the application of Koopman system estimation and control in multi-agent systems is provided in~\cite{tao2023koopmanI,zhao2023koopmanII}.

A critical aspect of multi-agent systems is formation control, where a group of robots must coordinate to maintain a specific spatial arrangement while in motion. 
However, real-world applications may introduce external environmental disturbances, posing challenges to the multi-agent system's robustness. 
The Koopman operator can help address these challenges. 
It has been employed to estimate the disturbed model of agents within a group, either through online adaptation~\cite{wang2022potentialonline} or offline training~\cite{wang2023uncertaintyoffline} methods. 
This estimation aids in managing environmental uncertainties, allowing the team to maintain a desired formation. 
%
Koopman operator theory has recently been used to address scenarios involving disconnection and signal recovery within the system~\cite{zhan2023signal}, by essentially 
aiding in recovering missed signals of the leader by capturing inherent features through linear motion evolution. 
It has also been applied for modeling high-dimensional biological or engineering swarms~\cite{hansen2022swarm}, thereby facilitating the learning of local interactions within homogeneous swarms based on observational data and enabling the generation of similar swarming behavior using the acquired model.



\section{Advanced Topics on Modeling and Control}\label{sec:advanced} 
We believe that several open challenges (discussed in Section~\ref{sec:Discussion} that follows) can be addressed by pushing forward the state of the theory. 
As such, we revisit the Koopman operator theory and present a series of advanced topics that delve deeper into important underlying theoretical aspects. 
We begin by extending our discussion to the continuous-time formulation of the Koopman operator, which is essential for systems governed by differential equations. While we have discussed practical implementations inspired by the Koopman operator for robotic systems with inputs earlier, here we provide a theoretically-grounded operator framework for incorporating control inputs, aiming to provide a fundamental and insightful perspective. 
Finally, we offer an in-depth discussion on the construction of lifting functions that enable accurate approximations of the Koopman operator.

\subsection{Koopman Operator Theory for Continuous-Time Systems}
In practice, the discrete-time formulation of the Koopman operator is most commonly used, as data are typically collected through sampling and employed for learning and adaptation. However, many real-world dynamical systems in physics, engineering, and robotics are inherently described by continuous-time dynamics. 
Therefore, in this section, we extend our introduction to include the continuous-time formulation of Koopman operator, providing a theoretical foundation for its application to continuous dynamical systems.

Let us now consider the system
\begin{align}\label{eq:continuous-time-system}
  \dot{x} = G(x), \quad x \in \Xc \subseteq \real^{N_x}\enspace,
\end{align}
where the map $G$ is continuously differentiable. 
For $t \in \realplus$, let $\Gc^t: \Xc \to \Xc$ be the associated flow map defined as
\begin{align}\label{eq:flow-map}
  \Gc^t(x(0)) := x(0)+ \int_{\tau = 0}^{t} G(x(\tau)) d\tau\enspace,
\end{align}
for all initial conditions $x(0) \in \Xc$. We assume the flow map \eqref{eq:flow-map} is well-defined for all $t \in \realplus$, i.e. it is a complete flow.

Similarly to the discrete-time case, consider the vector space $\Fc$ defined over the field $\cplx$, comprised of complex-valued functions defined on the domain $\Xc \subseteq \real^{N_x}$. Assume $\Fc$ is closed under composition with the flow map $\Gc^t$, for all $t \in \realplus$. Then, for each $t \in \realplus$, we can define a Koopman operator $\Kc^t: \Fc \to \Fc$, similarly to~\eqref{eq:Koopman-operator}, as 
\begin{align}\label{eq:Koopman-operator-flow}
  \Kc^t f = f \circ \Gc^t, \quad \forall f \in \Fc\enspace.
\end{align}

If $\Fc$ is a Banach space with norm $\| \cdot\|$ and the family of operators $\{\Kc^t\}_{t \in \realplus}$ is a strongly continuous semi-group,\footnote{~See for example~\cite[Chapter~1]{mauroy2020koopman} for specific choices of $G$ and $\Fc$ that turn $\{\Kc^t\}_{t \in \realplus}$ into a strongly continuous semi-group.} i.e. it satisfies
\begin{enumerate}
\item $\Kc^0 = \text{id}$\;,
\item $\Kc^{t_1 + t_2} = \Kc^{t_1} \Kc^{t_2}$, for all
  $t_1, t_2 \in \realplus$\;,
\item $\lim\limits_{t \searrow 0} \|\Kc^t f -f \| = 0$, for all
  $f \in \Fc$\;,
\end{enumerate}
where $\text{id}$ is the identity operator on $\Fc$, then we can define the infinitesimal generator $\Lc_{G}: \Fc \to \Fc$ of the semi-group $\{\Kc^t\}_{t \in \realplus}$ as
\begin{align}\label{eq:Koopman-generator}
  \Lc_{G} f := \lim\limits_{t \searrow 0} \frac{\Kc^t f - f}{t} = G
  \cdot \nabla f, \quad \forall f \in \Fc\;, 
\end{align}
where $\cdot$ and $\nabla$ represent the dot product and the gradient operator respectively. 
$\Lc_{G}$ is often referred to as the \emph{Koopman generator}.\footnote{~The definition of Koopman generator can be slightly relaxed to hold on a dense subset of $\Fc$~\cite[Chapter~1]{mauroy2020koopman}.} 

\subsection{Koopman Operators for Systems with Control Inputs} \label{subsec:advancedInputs}
Consider the control system 
\begin{equation}\label{eq:input-system}
    x_{t+1} = T_u(x_t,u_t)\;,
\end{equation}  
with state space $\Xc \subseteq \real^{N_x}$ and input space $\Uc \subseteq \real^{N_u}$. 
Dealing with systems in~\eqref{eq:input-system} from a Koopman operator perspective is significantly more difficult compared to systems without input. This challenge arises from the fundamental distinction between states and inputs: while the state is an intrinsic property of the system that evolves according to its internal dynamics, the input—especially in open-loop systems—is not known a priori and can significantly influence the system's evolution. In the following, we present two theoretical perspectives that aim to address this foundational issue and offer deeper insights into how inputs can be integrated into Koopman-based models.

\subsubsection{Considering All Infinite Input
  Sequences}
The work in~\cite{korda2018KoopmanMPC} tackles the aforementioned issue by considering the system's behavior over all possible input sequences. Formally, consider the space $l(\Uc)$ comprised of \emph{all possible infinite input sequences} $\{u_s(i)\}_{i=0}^\infty$ with $u_s(i) \in \Uc$ and let $S: l(\Uc) \to l(\Uc)$ denote the left-shift operator on $l(\Uc)$, mapping the sequence $\{u_s(i)\}_{i=0}^\infty$ to $\{u_s(i)\}_{i=1}^\infty$.  Now, consider the extended state $x_s := (x, u_s) \in \Xc \times l(\Uc)$ and define the dynamical system
\begin{align}\label{eq:infinite-input-dynamics}
  x_s^+ = (x,u_s)^+ = (T_u(x,u_s(0)), S u_s) =: L(x_s)\;, 
\end{align}
where $u_s(0)$ is the first element of $u_s$.  As in~\eqref{eq:Koopman-operator}, we can define a Koopman operator
$\Kc_L: \Hc \to \Hc$ for system $L$
in~\eqref{eq:infinite-input-dynamics} as
\begin{align}\label{eq:Koopman-infinite-sequence}
\Kc_L f = f \circ L, \; \forall f \in \Hc\;,
\end{align}
where the vector space $\Hc$ is defined over the field $\cplx$, consists of functions with domain $\Xc \times l(\Uc)$ and codomain $\cplx$, and is closed under composition with $L$, i.e.,
$g \circ L \in \Hc$ for all $g \in \Hc$.

Note that due to the dependency on infinite input sequences, working with operator~\eqref{eq:Koopman-infinite-sequence} is significantly more difficult than in the case for~\eqref{eq:Koopman-operator}, as it does not afford a direct way to find general finite-dimensional models for systems with access to finite input sequences.  However, since~\cite{korda2018KoopmanMPC} aims to use model predictive control, which requires \emph{relatively accurate short-term prediction}, it assumes the dynamics can be approximated by a lifted linear model (termed linear predictor) as
\begin{align}\label{eq:lifted-linear-control}
  z^+ \approx A z + Bu\;,
\end{align}
where $z$ is the lifted state starting from the initial condition
$z_0 = \Psib(x_0)$ (here $\Psib: \Xc \to \real^{N_\Psib}$ is the lifting map). 
One can estimate the matrices $A$ and $B$
in~\eqref{eq:lifted-linear-control} with an EDMD-like method. 
However, it is crucial to keep in mind that since the model does not consider infinite-input sequences, it does not generally capture all the information of the operator $\Kc_L$. 
Even if the dimension of the lifted state goes to infinity, one cannot generally conclude convergence of the lifted linear model trajectories to the trajectories of the nonlinear system; see also~\cite[\emph{Discussion~after~Corollary~1}]{korda2018KoopmanMPC}.

\subsubsection{Koopman Control Family}
An alternative Koopman-based approach to model the system does not rely on infinite input sequences and it is easier to use for
finite-dimensional approximations~\cite{MH-JC:24-auto}. The central idea is that one can represent the behavior of~\eqref{eq:input-system} by a family of systems in the form of~\eqref{eq:system} generated by setting the input in~\eqref{eq:input-system} to be a constant signal. 
Consider the family of constant input
dynamics $\{T_{\hat{u}}\}_{\hat{u} \in \Uc}$ defined by
\begin{align}\label{eq:constant-input-dynamis}
  x^+ = T_{\hat{u}} (x) := T_u(x, u \equiv \hat{u}), \; \hat{u} \in \Uc\;.
\end{align}
For any trajectory $\{x_i\}_{i=1}^{m+1}$ of~\eqref{eq:input-system} generated by input
sequence $\{u_i\}_{i=1}^m$ and initial condition $x_0$, one can write
\begin{align}\label{eq:constant-input-evolution}
  x_{m+1} = T_{u_m} \circ T_{u_{m-1}} \circ \cdots \circ T_{u_0}(x_0)\;,
\end{align}
Hence, the subsystems of the family $\{T_{\hat{u}}\}_{\hat{u} \in \Uc}$ completely capture the behavior of~\eqref{eq:input-system}. Moreover, we can use~\eqref{eq:Koopman-operator} to define Koopman operators for each subsystem. 
Consider the vector space $\Fc$ over field $\cplx$ comprised of complex-valued functions with domain $\Xc$ that is closed under composition with members of the family $\{T_{\hat{u}}\}_{\hat{u} \in \Uc}$. The \emph{Koopman Control Family} $\{\Kc_{\hat{u}}\}_{\hat{u} \in \Uc}$ is defined such that for all $\hat{u} \in \Uc$ we have
  \begin{align*}
    \Kc_{\hat{u}} g = g \circ T_{\hat{u}}, \; \forall g \in \Fc\;.
  \end{align*}

The Koopman Control Family can completely capture the evolution of functions in $\Fc$ over the trajectories of~\eqref{eq:input-system}. 
Given a trajectory $\{x_i\}_{i=1}^{m+1}$ of~\eqref{eq:input-system} generated by input sequence $\{u_i\}_{i=1}^m$ and initial condition $x_0$, one can write
\begin{align*}
g(x_{m+1}) = [\Kc_{u_0}\Kc_{u_1} \ldots \Kc_{u_m} g](x_0), \; \forall g \in \Fc\;.
\end{align*}

Similarly to the case without control inputs where Koopman-invariant subspaces lead to a finite-dimensional linear form, the models on \emph{common} invariant subspaces of the Koopman control family are \emph{all} in the following form (termed \emph{``input-state separable model''}, cf.~\cite[Theorem~4.3]{MH-JC:24-auto})
\begin{align*}
  \Phib(x^+) = \Phib \circ \Tc (x,u) = \Ac(u) \Phib(x)\;,
\end{align*}
where $\Phib: \Xc \to \cplx^{N_{\Phib}}$ is the lifting function and
$\Ac: \Uc \to \cplx^{N_{\Phib} \times N_{\Phib}}$ is a matrix-valued
function. Note that the input-state separable model is \emph{linear} in the lifted state but \emph{nonlinear} in the input. 
This is because, in an open-loop system, the input does not abide by predefined dynamics. 
Therefore, one cannot use the structure of the Koopman operator to represent the nonlinearity in the input as a linear operator. 
Interestingly, the popular lifted linear, bilinear, and linear switched models (e.g.,~\cite{SP-SK:19}) are all special cases of the input-state separable form (cf.~\cite[Lemmas~4.4-4.5]{MH-JC:24-auto}). 
In case there is no suitable common invariant subspace, one can approximate the action of the Koopman Control Family on any given subspace via orthogonal projections. 
See~\cite{MH-JC:24-auto} for theoretical analysis, data-driven learning methods, and accuracy bounds.

\subsection{Lifting Function Construction}\label{subsec:advancedLifting}
Here, we describe methods for the identification of dictionaries that lead to accurate Koopman approximations.
The accuracy of such approximations is directly related to the quality of the underlying subspace (the dictionary's span), specifically how close it is to being invariant under the Koopman operator. 

\subsubsection{Optimization-based Methods}
\begin{figure}[!t]
  \centering 
  {\includegraphics[width=.48\linewidth]{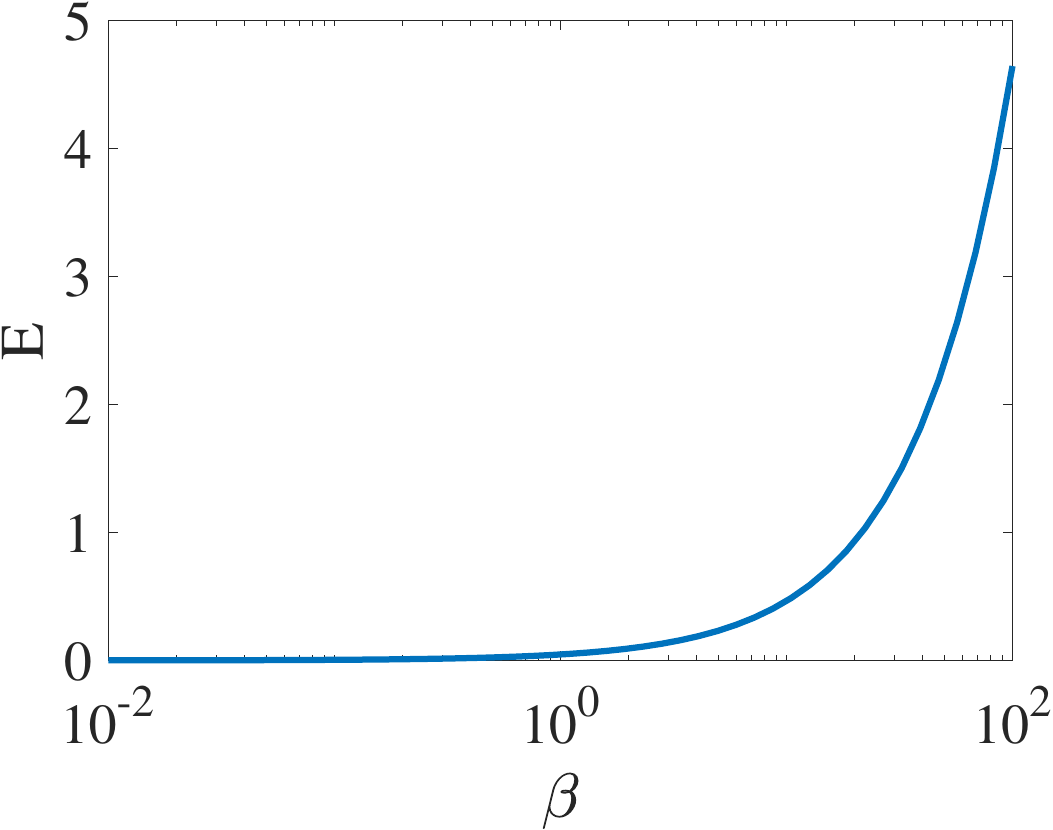}}{\includegraphics[width=.52\linewidth]{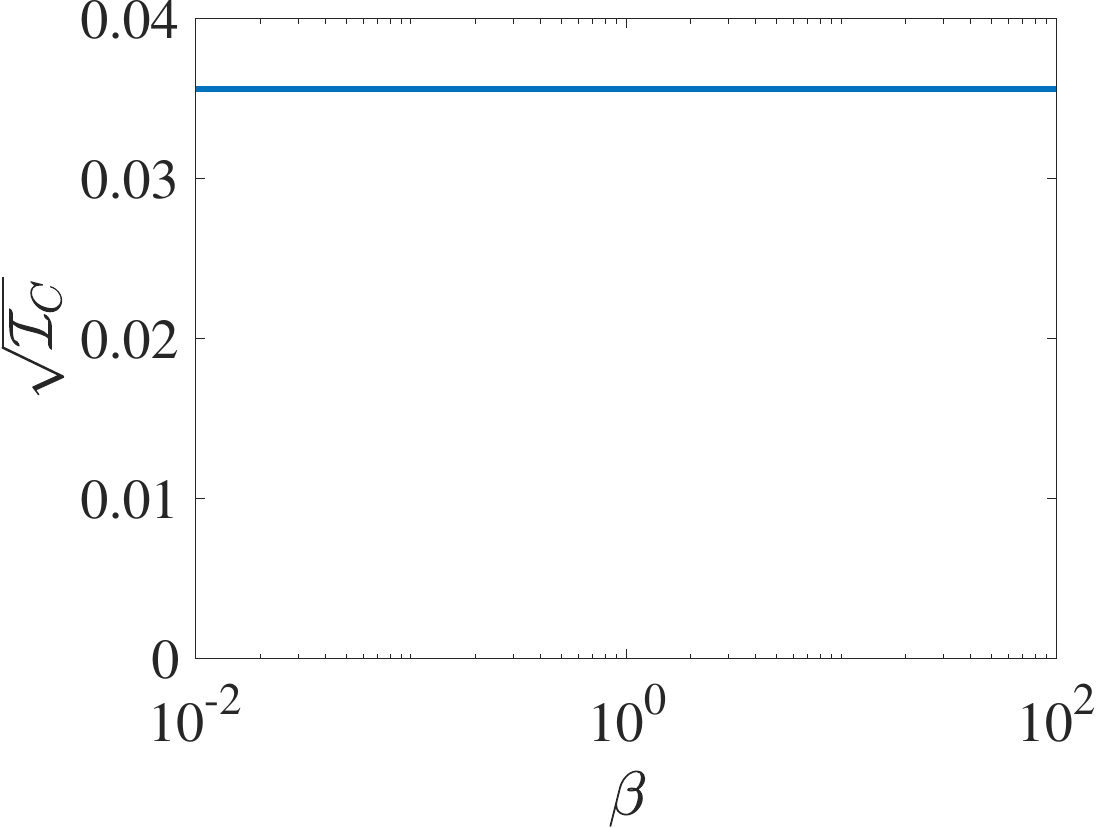}}
  \vspace{-6pt}
  \caption{\textbf{Left}: EDMD's residual error for linear system $x^+ = 0.6 x$ and the family of dictionaries $D_{\beta}(x) = [x, x + \beta \sin(x)]$ with $\beta \in [0.01, 100]$. Note that for all $\beta \in \real \setminus \{0\}$, all dictionaries span the same subspace $\Sc = \Span\{x,\sin(x)\}$. The residual error depends on the choice of basis for subspace $\Sc$. More importantly, $\Sc$ is not Koopman-invariant but the residual error can be arbitrarily close to zero depending on the basis. \textbf{Right}: The square root of consistency index for the system and family of dictionaries. Unlike EDMD's residual error, the consistency index does not depend on the choice of basis and accurately measures the approximation quality of the subspace.}\label{fig:not-invariant}
  \vspace*{-1.5ex}
\end{figure}

As introduced in the previous section, available data can be leveraged to learn a suitable dictionary of observables using neural networks or other parametric function families. This is typically achieved by minimizing the EDMD's residual error, formulated by combining equations~\eqref{eq:EDMD-optimization} and~\eqref{eq:EDMD-closed-form} as follows:
\begin{align}\label{eq:learning-loss-optim-reformulation}
\underset{ \Psib}{\text{minimize}} \| \Psib(Y) -
\Psib(Y) \Psib(X)^\dagger \ \Psib(X)\|_F;/.
\end{align}
Note that equation~\eqref{eq:learning-loss-optim-reformulation} is solved over dictionary $\Psib$ which belongs to a parametric family of functions; therefore, the optimization problem is generally non-convex, and finding globally optimal solutions is not guaranteed. More importantly, minimizing the EDMD's residual error does not necessarily lead to a close-to-invariant subspace. Figure~\ref{fig:not-invariant} (left) illustrates this point with an
example of a non-invariant subspace whose residual error can be made arbitrarily small depending on the choice of basis for the subspace. Therefore, optimization~\eqref{eq:learning-loss-optim-reformulation} may yield models unsuitable for long-term prediction.

The work in~\cite{MH-JC:23-csl} addresses this problem by providing an alternative loss function that captures the quality of the subspace. 
The notion of \emph{temporal forward-backward consistency index} (or \emph{consistency index} for brevity) is defined
as \footnote{~This definition is equivalent but different from the definition in~\cite{MH-JC:23-csl}.}
\begin{align}\label{eq:consistency-index}
  \ic(\Psib,X,Y) := \lambda_{\max} \big(I - K_F K_B \big)\;.
\end{align} 
$\lambda_{\max}(\cdot)$ denotes the maximum eigenvalue of its argument, and
\begin{align}\label{eq:forward-backward-EDMD}
  K_F = \Psib(Y) \Psib(X)^\dagger\;,
  \nonumber \\
  K_B = \Psib(X) \Psib(Y)^\dagger\;,
\end{align}
are the EDMD matrices applied forward and backward in time on the data set. 
The intuition behind this definition is that when the subspace is Koopman invariant, the forward and backward EDMD matrices are inverse of each other and therefore $\ic = 0$; otherwise, $\ic \neq 0$ and the larger $\ic$, the larger the deviation between forward and backward EDMD matrices from being inverse of each other.

The consistency index has several important properties. (1)~$\ic \in [0,1]$; (2)~Unlike the cost function in~\eqref{eq:learning-loss-optim-reformulation},
$\ic$ only depends on the subspace $\Span(\Psib)$ and not any specific basis (Fig.~\ref{fig:not-invariant} illustrates this); (3)~Under specific change of basis, $\ic$ can be viewed as the maximum eigenvalue of a positive semidefinite matrix ($I - K_F K_B$ is not generally symmetric) enabling the use of common
optimization solvers; (4)~Importantly, it provides a tight upper bound on the relative prediction error of EDMD on the entire subspace. 
Formally,

\begin{small}
\begin{align*}
  \sqrt{\ic(\Psib,X,Y)} = \max_{f \in \Span(\Psib), \|\Kc f
  \|_{L_2(\mu_X)} \neq 0} \frac{\| \Kc f - \Pf_{\Kc f}
  \|_{L_2(\mu_X)}}{\| \Kc f  \|_{L_2(\mu_X)}}, 
\end{align*} 
\end{small}

\noindent where $\Pf_{\Kc f}$ is the EDMD's predictor for $\Kc f$ in~\eqref{eq:EDMD-function-predictor} and the $L_2$ norm is calculated based on the empirical measure~\eqref{eq:empirical-measure}.

\begin{figure}[!t]
  \centering 
  \vspace{6pt}
  {\includegraphics[width=.7\linewidth]{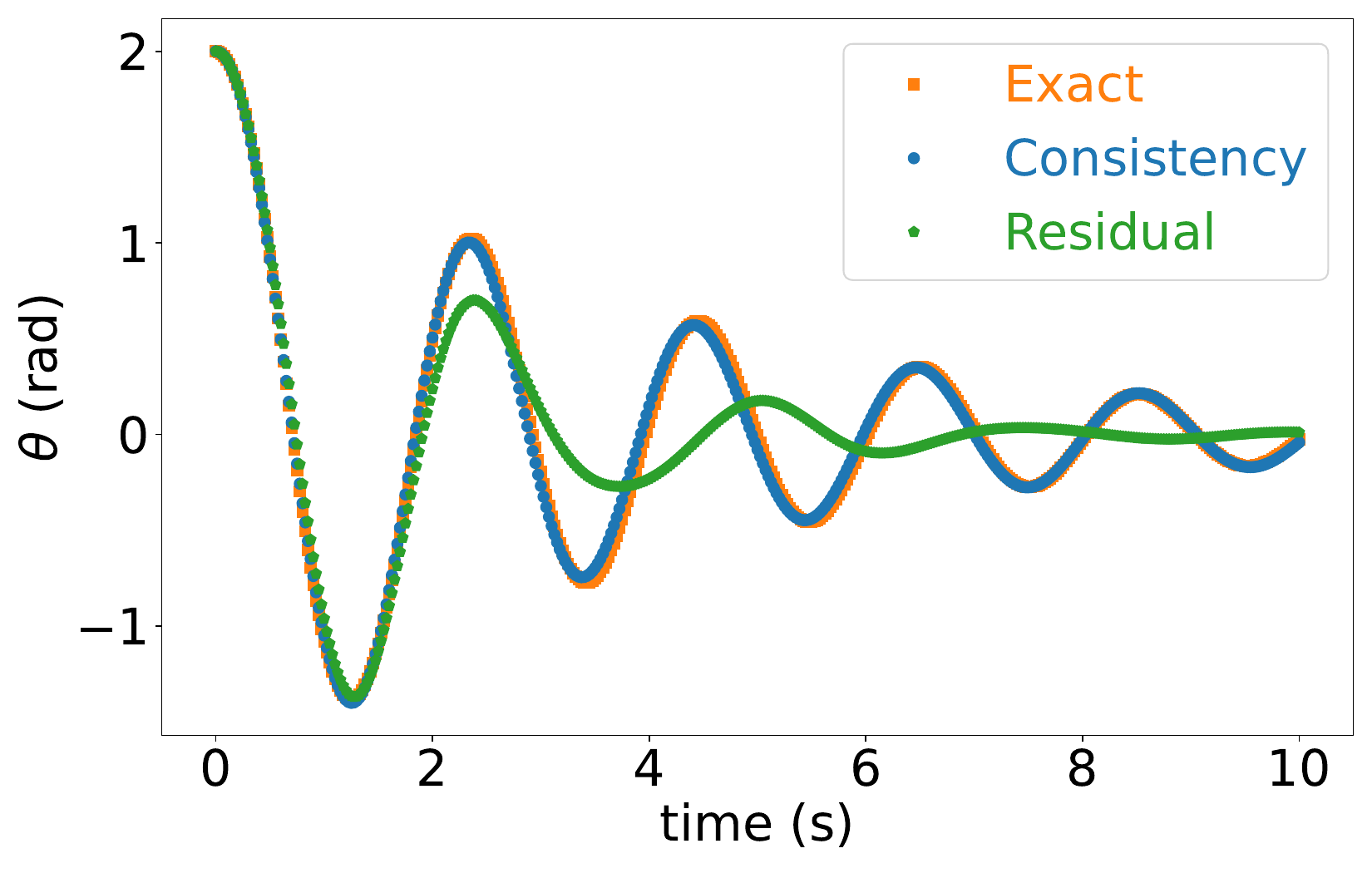}}
  \vspace{-6pt}
  \caption{We consider here the pendulum
    $[\dot{\theta}, \dot{\omega}] = [\omega, -9.81 \sin(\theta) - 0.1\omega]$ and a parametric family of dictionaries comprised of 5 functions in the form $\Psib(\theta,\omega) = [\theta, \omega, \text{NN}_1, \text{NN}_2,\text{NN}_3]$, where each NN is a feedforward neural network. The plot compares the prediction of the pendulum's angle evolution given linear predictors on the subspaces learned by minimizing the consistency index (equivalent to robust optimization~\eqref{eq:robust-learning}) and minimizing the residual error of EDMD (cf.~\eqref{eq:learning-loss-optim-reformulation}). The subspace learned by minimizing the consistency index is superior in long-term prediction. This is due to the fact that it accounts for all (uncountably many) members of the function space rather than only finitely many members considered in the residual error of EDMD.  }\label{fig:consistency-longterm-pendulum}
\end{figure}

Minimizing the consistency index is equivalent to the following robust minimax problem (where we have a closed-form expression for the max part)
\begin{align}\label{eq:robust-learning}
  \underset{ \Psib \in \text{PF}}{\text{minimize}} \max_{f \in
  \Span(\Psib), \| \Kc f  \|_{L_2(\mu_X)} \neq 0} \frac{\| \Kc f -
  \Pf_{\Kc f} \|_{L_2(\mu_X)}}{\| \Kc f  \|_{L_2(\mu_X)}}\;,\hspace{-3pt}
\end{align}
which minimizes the maximum EDMD function prediction error over the subspace. Note that, by minimizing the consistency index, one accounts for accurate prediction of all uncountably many members of the function space, as opposed to only finitely many functions in the optimization~\eqref{eq:learning-loss-optim-reformulation}. Therefore, by minimizing the consistency index, one can expect a more accurate
prediction, especially in the long term. Figure~\ref{fig:consistency-longterm-pendulum} illustrates this point in a nonlinear pendulum example.

\subsubsection{Algebraic Search for Koopman-Invariant Subspaces}
The optimization methods described above do not generally come with guarantees on the quality of the identified subspaces unless a global optimizer is found. 
However, this can be difficult since they often rely on objective functions that are non-convex and employ parametric families that lack desirable algebraic structures (e.g., neural networks). Interestingly, imposing the structure of a vector space on the parametric family allows one to effectively utilize the linearity of the Koopman operator for the identification of both exact and approximate eigenfunctions and invariant subspaces.

The work~\cite{MH-JC:22-tac} provides a data-driven necessary and almost surely sufficient condition to identify all Koopman eigenfunctions in an arbitrary finite-dimensional function space. 
This result directly relies on the eigendecomposition of forward and backward EDMD matrices (cf.~\eqref{eq:forward-backward-EDMD}). 
Moreover,~\cite{MH-JC:22-tac} provides an efficient and provably correct algebraic algorithm termed \emph{Symmetric Subspace Decomposition (SSD)} that finds the maximal Koopman-invariant subspace in arbitrary finite-dimensional function spaces (termed ``search space").~\cite{MH-JC:21-tcns} provides a parallel version of the SSD algorithm suitable for searching through high-dimensional spaces. 
Given as search space the finite-dimensional function space spanned by the dictionary $\Psib_s$, any basis $\Psib$ with elements in $\Span(\Psib_s)$ can be described using a matrix $C$ with full column rank as $\Psib^T(\cdot) = \Psib_s^T(\cdot) C$. 
If the subspace $\Span(\Psib)$ is invariant under the Koopman operator, this gets reflected in the data, $ \range(\Psib(X)^T) = \range(\Psib(Y)^T)$ (where $\range(\cdot)$ stands for the range space of its argument), which can also be written as
\begin{align}\label{eq:invariance-C}
  \range(\Psib_s (X)^T C ) = \range( \Psib_s (Y)^T C)\;.
\end{align}
Therefore, finding the largest invariant subspace within the finite-dimensional function space spanned by the dictionary $\Psib_s$ can be equivalently posed as finding the matrix $C$ with the maximum number of columns that satisfies~\eqref{eq:invariance-C}. 
The SSD algorithm and its parallel implementation build on this observation to devise an algebraic procedure to identify the largest invariant subspace contained in~$\Span(\Psib_s)$.

\begin{figure}[!t]
\vspace{6pt}
  \centering 
  {\includegraphics[width=.8\linewidth]{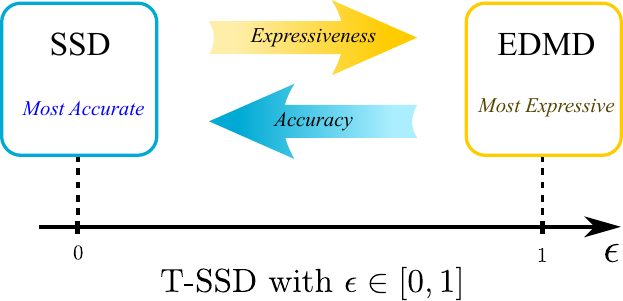}}
  \vspace{-6pt}
  \caption{The variable $\epsilon \in [0,1]$ in T-SSD sets the balance
    between the accuracy and expressiveness of the model. Both SSD and EDMD algorithms are special cases of T-SSD with $\epsilon =0$ and
    $\epsilon =1$, respectively. (Image taken from~\cite{MH-JC:23-auto}
    and is available under license (CC BY 4.0):    {https://creativecommons.org/licenses/by/4.0}.)
  }\label{fig:SSD-EDMD}
  \vspace{-6pt}
\end{figure}

\begin{figure}[!t]
  \centering 
  {\includegraphics[width=.95\linewidth]{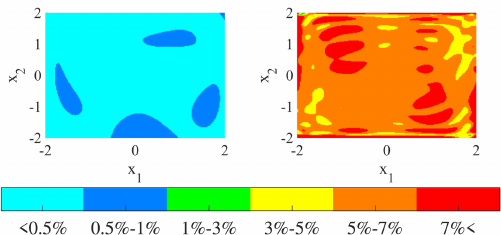}}
  \caption{Consider the Duffing System
    $[\dot{x}_1, \dot{x}_2] = [x_2, -0.5x_2 + x_1 (1-x_1)^2]$ over the
    state space $[-2,2]^2$ with the search space comprised of all polynomials up to degree 10. The right plot shows the relative EDMD dictionary prediction error for a normalized basis of the search space and the left plot shows the same error for the dictionary identified by T-SSD with $\epsilon = 0.02$. (Image taken from~\cite{MH-JC:23-auto} and is available under license (CC BY 4.0): {https://creativecommons.org/licenses/by/4.0}.)
  }\label{fig:TSSD_error_Duffing}
  \end{figure}

Exact Koopman-invariant subspaces capturing complete information about the dynamics are rare. 
A typical and useful approach is to allow for some error in the model to capture more (but inexact) information. 
However, it is crucial to characterize and tune the approximation accuracy. 
The \emph{Tunable Symmetric Subspace Decomposition (T-SSD)}~\cite{MH-JC:23-auto} is an algebraic search algorithm able to find a suitable subspace with any predetermined error level (specified by a tuning variable) on the function approximations.
Instead of requiring the equality in~\eqref{eq:invariance-C}, the T-SSD algorithm enforces the subspaces $\range(\Psib_s (X)^T C )$ and $\range( \Psib_s (Y)^T C)$ to be \emph{close}. Formally, this is captured by an accuracy parameter $\epsilon \in [0,1]$ that specifies the distance between the two subspaces. 
The T-SSD algorithm can be viewed as the generalization of the EDMD and SSD algorithms, cf. Fig.~\ref{fig:SSD-EDMD}. 
If we do not impose any accuracy level (allowing for up to a maximum 100\% prediction error by setting $\epsilon = 1$), T-SSD is equivalent to applying EDMD on the search space. 
On the other hand, if we specify zero-prediction error by setting $\epsilon = 0$, then T-SSD is equivalent to SSD and finds the maximal Koopman-invariant subspace of the search space with (almost surely) exact prediction on the entire subspace. 
In this sense, T-SSD balances the accuracy and the expressiveness (the dimension of the identified subspace) of the model based on the parameter~$\epsilon$. 
Figure~\ref{fig:TSSD_error_Duffing} shows the relative prediction error of the Duffing system over the search space of all polynomials up to degree 10 and the identified subspace by T-SSD given $\epsilon = 0.02$.
  
\section{Open Directions and Challenges}\label{sec:Discussion}
This review paper provided a comprehensive examination of the Koopman operator theory, a mathematical tool that can enable global linearization by elevating an original nonlinear system to a higher-dimensional linear space. 
We reviewed recent breakthroughs across a spectrum of robotic platforms, demonstrating how Koopman-based implementations are instrumental in addressing challenges related to robot runtime learning, modeling, and control. 
The paper meticulously described several foundational and more advanced yet key components regarding its deployment onto robots, including aspects related to data collection, the selection of lifting functions, and controller synthesis.

Despite the expanding use of the Koopman operator in modern robotics, there are still considerable challenges. 
The points that follow contain a curated summary of various discussions and insights derived from the authors' collective past and ongoing research. 
Certainly, the list is not exhaustive. 
However, we believe that it offers a solid starting point for future research directions catering to both practical implementations as well as further theoretical advancements.

\begin{enumerate}
    \item \textbf{Incorporating Constraints into Koopman Space:} 
    As in any optimization problem, incorporating the various (often conflicting) constraints appropriately is crucial. 
    While we have outlined some methods to do so, how to lift different types of constraints from the original space into the Koopman space requires more research. 
    We believe that exploiting the algebraic and geometric structure of the Koopman operator can play a key role.
    
    
    \item \textbf{Stochastic Simulation and Belief-Space Planning:} Real-world cases involve uncertainty, leading to the need for stochastic simulation and belief-space planning. 
    Handling multi-modal distributions and generating robust plans under uncertainty are vital research areas. Advancements in techniques like the stochastic Koopman operator~\cite{wanner2022SKO,takeishi2017SKO,han2021desko} can pave the way to incorporate stochasticity into Koopman-based control methods. 
    However, more work related to efficient implementation and the study of reliability and adaptability is needed.    
    \item \textbf{Sampling Rate Selection:} One of the ongoing inquiries surrounding the application of Koopman-based approaches in robotics pertains to the selection of an optimal sampling rate. 
    Particularly crucial for online learning, the sampling rate directly influences the quality of the derived Koopman operator. 
    Additionally, it governs the error bounds associated with predictions. 
    Selecting the rate for data collection and control signal execution will be vital to the performance of the robot.

    \item 
    \textbf{Extension to Fine/Dexterous Manipulation:} The emergence of Koopman-based approaches for dexterous manipulation marks a promising expansion of this methodology. 
    Contact-rich manipulation tasks, especially those involving fine-grained object interactions, non-prehensile control, and dexterous hand use, are becoming central challenges in robotics. 
    Koopman operator theory shows strong potential as a solution, offering a global linear representation that can handle the discontinuities introduced by contact dynamics. Moreover, its data-driven nature enables the capture of subtle, object-specific behaviors that are difficult to model analytically, while still maintaining a structure conducive to efficient optimization. 
    While this specific direction of research is still nascent, we believe it holds promise and merits more research.
    \item \textbf{Further Expansion on Soft Robotics:} One fundamental limitation of Koopman-based linearization is the need to lift the system into a high-dimensional space, which can become computationally expensive or even intractable for continuum or soft-bodied systems. Since soft robots naturally exhibit high degrees of freedom, achieving effective linearization often requires dimensionality reduction techniques or simplified parametric representations. However, these approximations inevitably introduce biases and limit generalizability, particularly in contact-rich scenarios or tasks involving large deformations. How to choose the most appropriate description of a soft robot for compatibility with Koopman methods remains an open area of research.
    \item \textbf{Extension to Hybrid Systems}: We have shown how Koopman operator theory can handle both discrete-time and continuous-time dynamical systems independently.
    Yet, the case of hybrid dynamical systems, which encapsulates many mechatronic and robotic systems, is more subtle and has received less attention to date. 
    A very recent effort has shown that, under certain conditions, unforced heterogeneous systems (i.e. a class of hybrid systems) can still be lifted into a higher-dimensional linear system via Koopman-based method~      \cite{asada2023global}. 
    However, extending to systems with inputs and systems whose evolution is governed by broader hybrid dynamics remains open.
    \item \textbf{Uncertainty in Lifted Features:} Lastly yet importantly, the assumption made in prior work regarding the zero-mean, Gaussian uncertainty model of the approximate Koopman operator should be revisited as well. 
    Recent advancements have demonstrated that one can still operate and plan controllers that are robust to this specific choice of uncertainty model, as done in~\cite{sinha2020robust, mamakoukas2022robust}. 
    Indeed, it is not clear if the nature of uncertainty in observables, especially with the addition of control input, can be accurately modeled as a normal Gaussian distribution. 
    As an example, one might have a system where the states $x_t$ are well modeled as Gaussian (in this case the underlying dynamics might also be linear). 
    It is not clear whether, and under which conditions, Gaussian distributions pushed into the observable lifted space should form Gaussians. 
    Thus, finding cases where lifting observables also preserves some statistical structure is a key direction of future research. 
    \item \textbf{Study of the Underlying Optimization Problems:}
    We have seen that the lifted system in the Koopman space is described by a linear (most often) or bilinear model, and that representing the (higher-dimensional) Koopman space observables with a (bi-)linear model does not conflict with the original (lower-dimensional) dynamics being nonlinear. While this property facilitates runtime adaptation, it may create challenges when solving the underlying optimization problems for control. Unless for special cases, the cost function can often depend on the lifted states. 
    This can create challenges, e.g., in assessing the recursive feasibility of Koopman-NMPC methods. 
    Although some challenges can be mitigated by enforcing a more careful implementation,
    how to design appropriate cost functions that can generalize across robotic systems remains an open problem. 
\end{enumerate}

\section{Conclusion}\label{sec:Conclusion}
In conclusion, the application of Koopman operator theory in robotics represents a transformative shift in how we approach the modeling, control, and optimization of complex robotic systems in support of runtime learning. 
By leveraging the power of a linear representation for inherently nonlinear dynamics, Koopman-based methods provide a robust and computationally efficient framework that addresses many of the limitations of traditional approaches. 
The theoretical advancements and practical implementations discussed in this review demonstrate the versatility and efficacy of Koopman operators across a wide array of robotic domains, including aerial, legged, wheeled, underwater, soft, and manipulator robots. 
We hope this survey can serve as an inviting gateway for newcomers, helping readers quickly build intuition for the theoretical underpinnings of the Koopman operator while also discovering its exciting applications across modern robotics. By demystifying the concepts and showcasing real-world use cases, we aim to spark curiosity, lower the barrier to entry, and empower more researchers and practitioners to explore, experiment, and advance this powerful framework. 
We believe Koopman-based methods hold tremendous potential, and we are excited to see how the robotics community will build on this foundation to drive future innovation.

\balance
\bibliographystyle{IEEEtran}
\bibliography{IEEEabrv,IEEEref}

\end{document}